%% file: main.tex
\documentclass[runningheads]{llncs}

\usepackage{eccv}

\input{preamble}

\usepackage{colortbl}
\usepackage{eccvabbrv}
\usepackage{wrapfig}
\usepackage{graphicx}
\usepackage{booktabs}

\usepackage[accsupp]{axessibility}  %

\usepackage[pagebackref,breaklinks,colorlinks,citecolor=eccvblue]{hyperref}

\usepackage{orcidlink}

\usepackage[sort,numbers]{natbib}

\begin{document}

\title{Similarity of Neural Architectures\\using Adversarial Attack Transferability}

\author{Jaehui Hwang$^{1,2,\dagger}$ \quad Dongyoon Han$^3$ \quad Byeongho Heo$^3$ \quad Song Park$^3$\\
{Sanghyuk Chun$^{3,*}$ \quad Jong-Seok Lee$^{1,2,*}$}}

\authorrunning{J. Hwang et al.}

\institute{{\small $^1$ School of Integrated Technology, Yonsei University\\
$^2$BK21 Graduate Program in Intelligent Semiconductor Technology, Yonsei University \\
$^{3}$ NAVER AI Lab} \\ \, \\
$^{\dagger}$ {\footnotesize Works done during an internship at NAVER AI Lab.}
$^{*}$ {\footnotesize Corresponding authors}
}

\maketitle

\begin{abstract}
In recent years, many deep neural architectures have been developed for image classification. Whether they are similar or dissimilar and what factors contribute to their (dis)similarities remains curious. To address this question, we aim to design a quantitative and scalable similarity measure between neural architectures. We propose Similarity by Attack Transferability (SAT) from the observation that adversarial attack transferability contains information related to input gradients and decision boundaries widely used to understand model behaviors. We conduct a large-scale analysis on 69 state-of-the-art ImageNet classifiers using our SAT to answer the question. In addition, we provide interesting insights into ML applications using multiple models, such as model ensemble and knowledge distillation. Our results show that using diverse neural architectures with distinct components can benefit such scenarios.
\end{abstract}

\begin{wrapfigure}{r}{.45\linewidth}
\centering
\vspace{-2.5em}
\includegraphics[width=\linewidth]{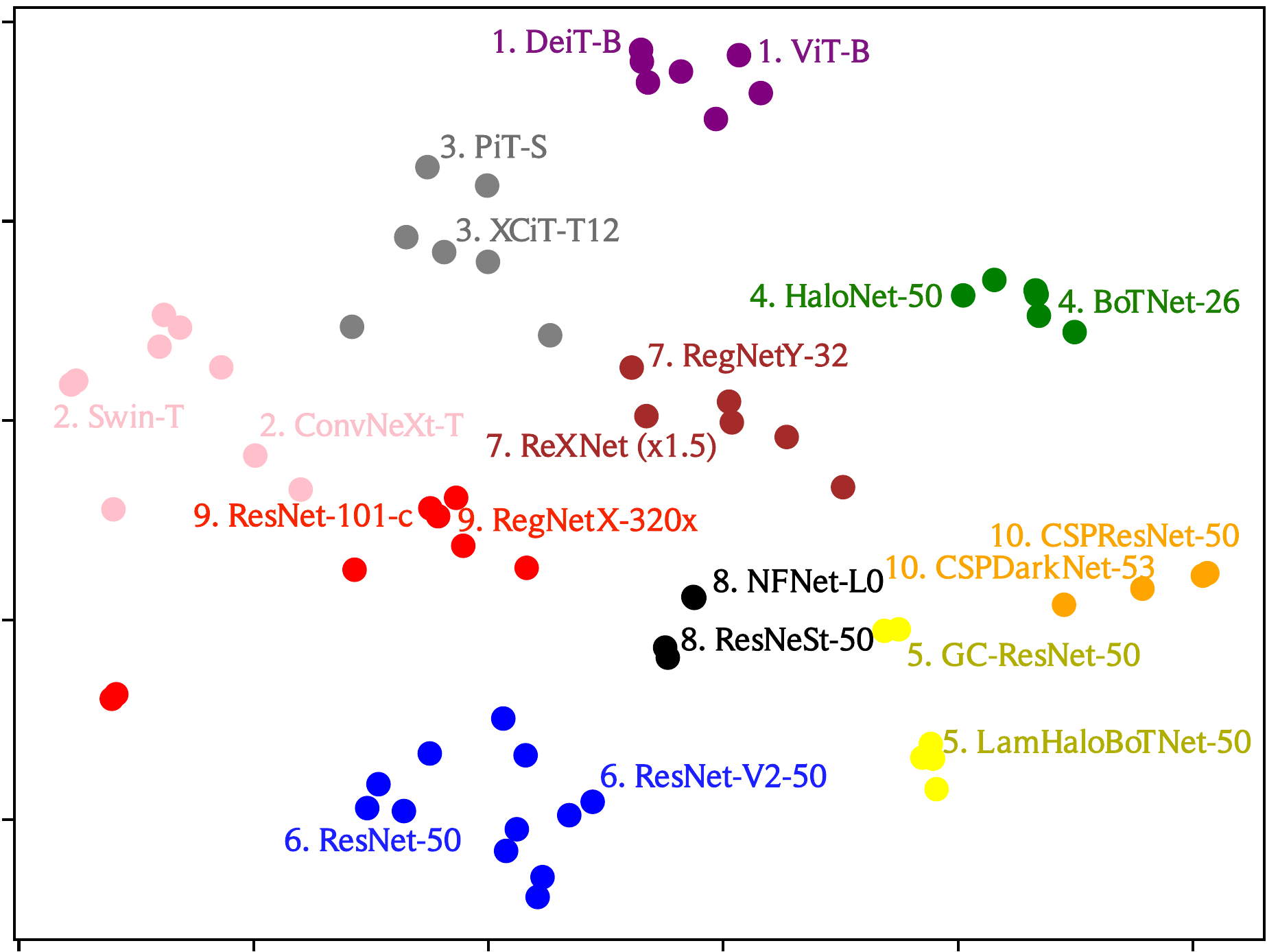}
\caption{\small {\bf t-SNE plot showing 10 clusters of 69 neural networks using our similarity function, \ours.}
}
\label{fig:tsne_teaser}
\vspace{-1em}
\end{wrapfigure}
\section{Introduction}
The advances in deep neural networks (DNN) architecture design have taken a key role in their success
by making the learning process easier (\eg, normalization \cite{batchnorm, wu2018groupnorm, ba2016layernorm} or skip connection~\cite{resnet}), enforcing human inductive bias \cite{krizhevsky2012alexnet}, or increasing model capability with the self-attention mechanism \cite{vaswani2017attention}. With different architectural components containing architectural design principles and elements, a number of different neural architectures have been proposed. They have different accuracies, but several researches have pointed out that their predictions are not significantly different \cite{trivial, mania2019model, geirhos2020beyond}.

By this, \textit{can we say that recently developed DNN models with different architectural components are similar or the same?}
The answer is \textit{no}. It is because a model prediction is not the only characteristic to compare their similarities.
Existing studies have found differences by focusing on different features, such as layer-by-layer network component \cite{kornblith2019cka,raghu2021vision}, a high-level understanding by visualization of loss surface \cite{dinh2017sharp}, input gradient \cite{GuidedBackprop,SmoothGrad}, and decision boundary \cite{somepalli2022reproducibility}.
Researchers could understand the similarity between models through these trials; however, the similarity comparison methods from previous studies are insufficient for facilitating comprehensive studies because they do not satisfy two criteria that practical metrics should meet:
(1) providing a quantitative similarity score and (2) being compatible with different base architectures (\eg, CNN and Transformer).
Recently, \citet{tramer2017space} and \citet{somepalli2022reproducibility} suggested a quantitative similarity metric based on measuring differences in decision boundaries.
However, these methods have limitations due to the non-tractable decision boundaries and limited computations as shown in \cref{sec:similarity}.

We propose a quantitative similarity that is scalable and easily applicable to diverse architectures, named \oursfull.
We focus on adversarial attack transferability (AT), which indicates how generated adversarial perturbation is transferable between two different architectures.
It is widely studied that the vulnerability of DNNs depends on their own architectural property or how models capture the features from inputs, such as the usage of self-attention \cite{patchfool}, the stem layer \cite{hwang2021just}, and the dependency on high or low-frequency components of input \cite{bai2022improving, kim2022analyzing}.
Thus, if two different models are similar, the AT between the models is high because they share similar vulnerability \cite{rezaei2019target}.
Furthermore, AT can be a reliable approximation for comparing the input gradients \cite{madry2017pgd}, decision boundary \cite{karimi2020decision}, and loss landscape \cite{demontis2019adversarial}. All of them are widely-used frameworks to understand model behavior and differences between models and used to measure the similarity of models in previous works \cite{GuidedBackprop,SmoothGrad,IntegratedGradients,bansal2020sam,choe2020wsolevalextension, somepalli2022reproducibility, dinh2017sharp,li2018visualizing,tramer2017space,waseda2023closer}; namely, \ours can capture various model properties.

We quantitatively measure pairwise \ourss of 69 different ImageNet-trained neural architectures from \cite{rw2019timm}. We analyze what components among 13 architectural components (\eg, normalization, activation, \ldots) that consist of neural architectures largely affect model diversity.
Furthermore, we observe relationships between SAT and practical applications, such as ensemble and distillation.

\section{Related Work}

\textbf{Similarity between DNNs} has been actively explored recently.
Several studies focused on comparing intermediate features to understand the behavior of DNNs. \citet{raghu2021vision} observed the difference between layers, training methods, and architectures (\eg, CNN and ViT) based on \textbf{layer-by-layer comparison} \cite{kornblith2019cka}. Some studies have focused on \textbf{loss landscapes} by visualizing the loss of models on the parameter space \cite{dinh2017sharp,li2018visualizing,park2022vision}. Although these methods show a visual inspection, they cannot support quantitative measurements. On the other hand, our goal is to support a quantitative similarity by \ours.

Another line of research has been focused on \textbf{prediction-based statistics}, \eg, comparing wrong and correct predictions \cite{kuncheva2003measures,geirhos2018generalisation,geirhos2020beyond,scimeca2022wcst-ml}. However, as recent complex DNNs are getting almost perfect, just focusing on prediction values can be misleading;
\citet{trivial} observed that recent DNNs show highly similar predictions. In this case, prediction-based methods will be no more informative.
Meanwhile, our \ours can provide meaningful findings for 69 recent NNs.

\textbf{Input gradient} is another popular framework to understand model behavior by observing how a model will change predictions by local pixel changes \cite{FirstDNNInputGradient,GuidedBackprop,SmoothGrad,IntegratedGradients,bansal2020sam}.
If two models are similar, their input gradients will also be similar. These methods are computationally efficient, and no additional training is required; they can provide a visual understanding of the given input. 
However, input gradients are inherently noisy; thus, these methods will need additional pre-processing, such as smoothing, for a stable computation \cite{choe2020wsolevalextension}. Also, these methods usually measure how the input gradient matches the actual foreground, \ie, we need ground-truth foreground masks for measuring such scores. On the contrary, \ours needs no additional pre-processing and mask annotations.

\textbf{Comparing the decision boundaries} will provide a high-level understanding of how models behave differently for input changes and how models extract features from complicated data dimensions. Recent works \cite{tramer2017space, waseda2023closer} suggested measuring similarity by comparing distances between predictions and decision boundaries. Meanwhile, \citet{somepalli2022reproducibility} analyzed models by comparing their decision boundaries on the on-manifold plane constructed by three random images.
However, these approaches suffer from inaccurate approximation, non-tractable decision boundaries, and finite pairs of inputs and predictions.

Finally, different behaviors of CNNs and Transformers have been studied in specific tasks, such as robustness \cite{bai2021transformers,naseer2021intriguing}, layer-by-layer comparison \cite{raghu2021vision,park2022vision} or decision-making process \cite{jiang2024comparing}. Our work aims to quantify the similarity between general NNs, not only focusing on limited groups of architecture.

\section{\oursfull}
\label{sec:similarity}

Here, we propose a quantitative similarity between two architectures using adversarial attack transferability, which indicates whether an adversarial sample from a model can fool another model.
The concept of adversarial attack has effectively pointed out the vulnerabilities of DNNs by input gradient \cite{szegedy2013intriguing, goodfellow2014explaining, madry2017pgd}.

Interestingly, these vulnerabilities have been observed to be intricately linked to architectural properties.
For example, \citet{patchfool} demonstrated the effect of the attention modules in architecture on attack success rate.
\citet{hwang2021just} analyzed that the stem layer structure causes models to have different adversarial vulnerable points in the input space, \eg, video models periodically have vulnerable frames, such as every four frames. Namely, an adversarial sample to a model highly depends on the inherent architectural property of the model.

Another perspective emphasized the dissimilarities in dependencies on high-frequency and low-frequency components between CNN-based and transformer-based models, showing different vulnerabilities to different adversarial attacks \cite{bai2022improving, kim2022analyzing}. Different architectural choices behave as different frequency filters (\eg, the self-attention works as a low-pass filter, while the convolution works as a high-pass filter) \cite{park2022vision}; thus, we can expect that the different architectural component choices will affect the model vulnerability, \eg, vulnerability to high-frequency perturbations. If we can measure how the adversarial vulnerabilities of the models are different, we also can measure how the networks are dissimilar.

To measure how model vulnerabilities differ, we employ \textbf{adversarial attack transferability} (AT), where it indicates whether an adversarial sample from a model can fool another model. If two models are more similar, their AT gets higher \cite{rezaei2019target, demontis2019adversarial, modeldiff}. On the other hand, because the adversarial attack targets vulnerable points varying by architectural components of DNNs \cite{hwang2019prm, patchfool, hwang2021just, kim2022analyzing}, if two different models are dissimilar, the AT between them gets lower. Furthermore, attack transferability can be a good approximation for measuring the differences in input gradients \cite{madry2017pgd}, decision boundaries \cite{karimi2020decision}, and loss landscape \cite{demontis2019adversarial}, where they are widely used techniques for understanding model behavior and similarity between models as discussed in the related work section. While previous approaches are limited to non-quantitative analysis, inherent noisy property, and computational costs, adversarial transferability can provide quantitative measures with low variances and low computational costs.

We propose a new similarity function that utilizes attack transferability, named \textbf{\oursfull}, providing a reliable, easy-to-conduct, and scalable method for measuring the similarity between neural architectures.
Formally, we generate adversarial samples $x_A$ and $x_B$ of model $A$ and $B$ for the given input $x$. Then, we measure the accuracy of model $A$ using the adversarial sample for model $B$ (called $\text{acc}_{B \rightarrow A}$). If $A$ and $B$ are the same, then $\text{acc}_{B \rightarrow A}$ will be zero if the adversary can fool model B perfectly. On the other hand, if the input gradients of $A$ and $B$ differ significantly, then the performance drop will be neglectable because the adversarial sample is almost similar to the original image (\ie, $\|x - x_B\| \leq \varepsilon$). Let $X_{AB}$ be the set of inputs where both $A$ and $B$ predict correctly, $y$ be the ground truth label, and $\mathbb I (\cdot)$ be the indicator function. We measure \ours between two different models by:
\begin{equation}
\text{SAT}(A,B) = \log \big[\max \big\{ \varepsilon_s, 100 \times
\frac{1}{2|X_{AB}|}
\sum_{x \in X_{AB}} \left \{ \mathbb {I} ({A(x_B)} \neq y) + \mathbb {I} ({B(x_A)} \neq y) \right \} \big \} \big],
\label{eq:score}
\end{equation}
\noindent where $\varepsilon_s$ is a small scalar value.
If $A=B$ and we have an oracle adversary, then $\text{SAT}(A,A) = \log 100$. In practice, a strong adversary (\eg, PGD \cite{madry2017pgd} or AutoAttack \cite{autoattack}) can easily achieve a nearly-zero accuracy if a model is not trained by an adversarial attack-aware strategy \cite{madry2017pgd, cohen2019certified}. Meanwhile, if the adversarial attacks on $A$ are not transferable to $B$ and vice versa, then $\text{SAT}(A,B) = \log \varepsilon_s$.

Ideally, we aim to define a similarity $d$ between two models with the following properties: (1) $n = \arg\min_m d(n, m)$, (2) $d(n,m) = d(m,n)$ and (3) $d(n, m) > d(n, n)$ if $n \neq m$.
If the adversary is perfect, then $\text{acc}_{A \rightarrow A}$ will be zero, and it will be the minimum because accuracy is non-negative. ``$\text{acc}_{A \rightarrow B} + \text{acc}_{B \rightarrow A}$'' is symmetric thereby SAT is symmetric. Finally, SAT satisfies $d(n, m) \geq d(n, n)$ if $n \neq m$ where it is a weaker condition than (3). 

\begin{figure}[t]
    \centering
    \includegraphics[width=.79\linewidth]{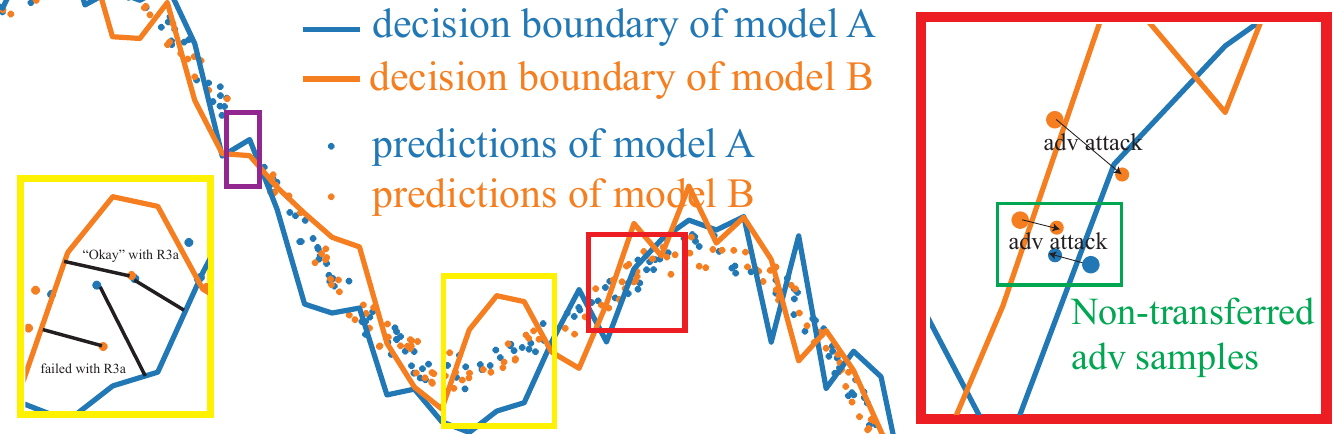}
    \caption{\textbf{How \ours works?} Conceptual figure to understand \ours by the lens of the decision boundary. Each line denotes the decision boundary of a binary classification model, and each dot denotes individual prediction for given inputs.}
    \label{fig:thm}
\end{figure}

\paragraph{Comparison with other methods.} Here, we compare \ours with prediction-based measurements \cite{kuncheva2003measures,geirhos2018generalisation,geirhos2020beyond,scimeca2022wcst-ml} and similarity measurements by comparing decision boundaries (\citet{tramer2017space} and \citet{ somepalli2022reproducibility}). We first define two binary classifiers $f$ and $g$ and their predicted values $f_p(x)$ and $g_p(x)$ for input $x$ (See \cref{fig:thm}). $f$ classifies $x$ as positive if $f_p(x) > f_d(x)$ where $f_d(x)$ is a decision boundary of $f$. We aim to measure the difference between decision boundaries, namely $\int_x |f_d(x) - g_d(x)| dx$ to measure differences between models.
However, DNNs have a non-tractable decision boundary function, thus, $f_d$ and $g_d$ are not tractable.
Furthermore, the space of $x$ is too large to compute explicitly. Instead, we may assume that we only have finite and sparingly sampled $x$. 

In this scenario, we can choose three strategies. First, we can count the number of samples whose predicted labels are different for given $x$, which is \textit{prediction-based measurements} or \citet{somepalli2022reproducibility}.
As we assumed sparsity of $x$, this approach cannot measure the area of uncovered $x$ domain, hence, its approximation will be incorrect (purple box in \cref{fig:thm}) or needs too many perturbations to search uncovered $x$. In \cref{subsec:appendix_empricial_comparison_somepalli}, we empirically show that \citet{somepalli2022reproducibility} suffers from the high variance even with a large number of samples while \ours shows a low variance with a small number of samples.

Second, we can measure the minimum distance between $f_p(x)$ and $f_d$ as \citet{tramer2017space}. This only measures the distance to its closest decision boundary without considering the other model. The yellow box of \cref{fig:thm} shows if two predictions are similar at $x$, it would compute an approximation of $|f_d(x) - g_d(x)|$ for $x$. However, if two predictions are different, it will compute a wrong approximation. Moreover, in practice, searching $\epsilon$ is unstable and expensive.

Lastly, we can count the number of non-transferred adversarial samples (red box in \cref{fig:thm}), which is \textit{our method, \ours}. If we have an oracle attack method that exactly moves the point right beyond the decision boundary, our SAT will measure the $\ell_0$ approximation of $\min(|f_d(x) - g_d(x)|, \epsilon)$ for given $x$. Namely, SAT can measure whether two decision boundaries are different by more than $\epsilon$ for each $x$. If we assume that the difference between decision boundaries is not significantly large and $\epsilon$ is properly chosen, SAT will compute an approximated decision boundary difference. We also compare \ours and other methods from the viewpoint of stability and practical usability in \cref{subsec:ensemble} and \cref{sec:appendix_empricial_comparison}.

\paragraph{Discussions.}
In practice, we do not have an oracle attack method. Instead, we employ the PGD attack \cite{madry2017pgd} as the adversarial attack method.
In \cref{subsec:appendix_attack_robustness}, we investigate the robustness of \ours to the choice of the attack methods. In summary, \ours measured by PGD shows a high correlation with \ours measured by various attacks, \eg, AutoAttack \cite{autoattack}, attacks designed for enhancing attack transferability, such as MIFGSM \cite{MIFGSM} and VMIFGSM \cite{VMIFGSM}, low-frequency targeted attacks, such as low-frequency PGD \cite{lowfpgd}, method-specific attacks, such as PatchPool \cite{patchfool}, or generative model-based attacks, such as BIA \cite{BIA}.

Also, \ours assumes an optimal attack with proper $\epsilon$. However, this assumption can be broken under the adversarial training setting when we use a practical attacker. Also, as shown by \citet{tsipras2018robustness} and \citet{ilyas2019adversarial}, adversarial training will lead to a different decision boundary from the original model. In \cref{subsec:appendix_impact_of_adv_tr}, we empirically investigate the effect of adversarial training to \ours. We observe that different adversarial training methods make as a difference as different training techniques, which we will discuss in \cref{subsec:trainging-strategy-similarity}.

\paragraph{Analyzing 69 models.}
Now, we analyze 69 recent ImageNet classifiers using \ours by focusing on two questions. (1) Which network component contributes to the diversity between models? (2) Why do we need to develop various neural architectures?
The full list of the architectures can be found in \cref{sec:appendix-models}.
We use the PGD attack \cite{madry2017pgd} for the adversary. We set the iteration to 50, the learning rate to 0.1, and $\varepsilon$ to $8/255$. As we discussed earlier, we show that \ours is robust to the choice of the adversarial attack method. We select 69 neural architectures trained on ImageNet \cite{imagenet} from the PyTorch Image Models library \cite{rw2019timm}.
To reduce the unexpected effect of a significant accuracy gap, the chosen model candidates are limited to the models whose top-1 accuracy is between 79\% and 83\%. We also ignore the models with unusual training techniques, such as training on extra training datasets, using a small or large input resolution (\eg, less than 200 or larger than 300), or knowledge distillation. When $A$ and $B$ take different input resolutions, then we resize the attacked image from the source network for the target network. We also sub-sample 10\% ImageNet validation images (\ie, 5,000 images) to measure the similarity. This strategy makes our similarity score more computationally efficient.

\section{Model Analysis by Network Similarity}

\subsection{Which Architectural Component Causes the Difference?}
\label{subsec:model-analysis}

\paragraph{Settings.}
We list 13 key architecture components: normalization (\eg, BN \cite{batchnorm} and LN \cite{ba2016layernorm}), activations (\eg, ReLU \cite{krizhevsky2012alexnet} and GeLU \cite{ramachandran2017searching}), the existence of depthwise convolution, or stem layer (\eg, 7$\times$7 conv, 3$\times$3 conv, or 16$\times$16 conv with stride 16 -- a.k.a. \textit{``patchify''} stem \cite{convnext}). 
The list of the entire components is shown in the Appendix. 
We then convert each architecture as a feature vector based on the listed sub-modules. For example, we convert ResNet as $f_\text{ResNet} = [\text{Base arch}=\text{CNN}, \text{Norm}=\text{BN},\text{Activation}=\text{ReLU}, \ldots]$.
The full list of components of 69 architectures can be found in \cref{sec:appendix-models}.

\begin{figure}[t]
\begin{minipage}[t]{0.6\linewidth}
    \centering
    \includegraphics[width=0.9\linewidth]{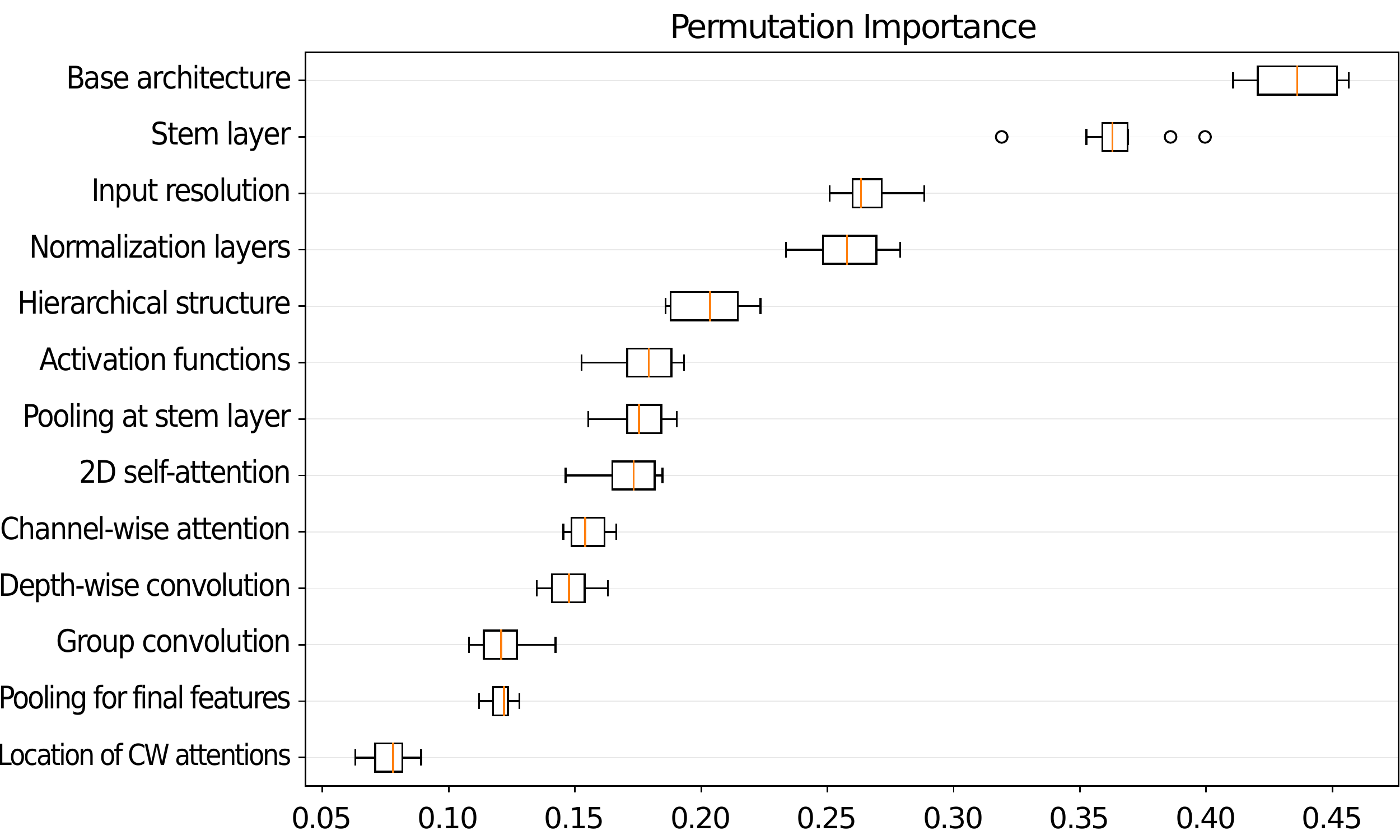}%
    \caption{\small {\bf Importance of architectural components to network similarity.} 13 components are sorted by the contribution to the similarities. The larger feature importance means the component contributes more to the network similarity.}
    \label{fig:model-analysis}
\end{minipage}\hspace{0.05\linewidth}
\begin{minipage}[t]{0.35\linewidth}
    \centering
    \includegraphics[width=0.9\linewidth]{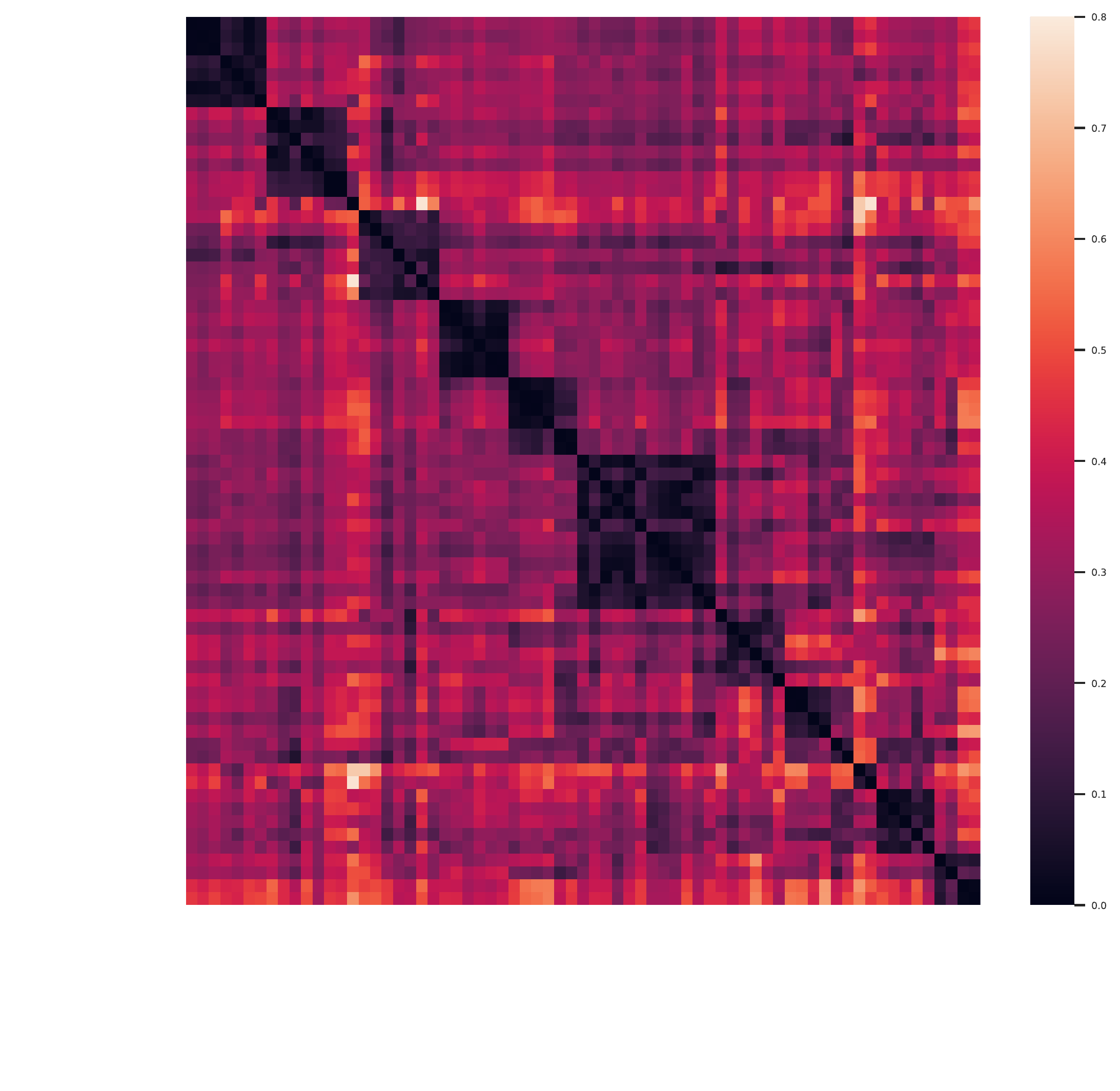}
    \caption{\small {\bf Pairwise distances of spectral features.} Rows and columns are sorted by the clustering index. More details are described in \cref{subsec:appendix-spectral-clustering}.}
    \label{fig:similarity-different-vs-same}
\end{minipage}\hfill
\end{figure}

\paragraph{Feature important analysis.}
Now, we measure the feature importance by fitting a gradient boosting regressor \cite{friedman2001greedy} on the feature difference (\eg, $f_\text{ResNet-50} - f_\text{DeiT-base}$) measured by Hamming distance and the corresponding similarity. The details of the regressor are described in Appendix. We use the permutation importance \cite{breiman2001random} that indicates how the trained regression model changes the prediction according to randomly changing each feature. The feature importance of each architectural component is shown in \cref{fig:model-analysis}. We first observe that the choice of base architecture (\eg, CNN \cite{krizhevsky2012alexnet}, Transformer \cite{vaswani2017attention}, and MLP-Mixer \cite{tolstikhin2021mlp}) contributes to the similarity most significantly. \cref{fig:model-analysis} also shows that the design choice of the input layer (\ie, stem layer design choice or input resolution) affects the similarity as much as the choice of basic components such as normalization layers, activation functions, and the existence of attention layers. On the other hand, we observe that the modified efficiency-aware convolution operations, such as depth-wise convolution \cite{xception}, are ineffective for diversity.

\begin{table}[t]
\renewcommand{\thefootnote}{\fnsymbol{footnote}}
\small
\centering
\renewcommand{\arraystretch}{0.9}
\caption{\small {\bf Clusters by SAT.} All the architectures here are denoted by the aliases defined in their respective papers. We show the top-5 keywords for each cluster based on TF-IDF. InRes, SA, and CWA denote input resolution, self-attention, and channel-wise attention, respectively. The customized model details are described in the footnote$^\dagger$.}
\label{tab:clustering-results}
\resizebox{\textwidth}{!} {
\begin{tabular}{@{}cll@{}}	
\toprule
No. & Top-5 Keywords & Architecture\\
\midrule
\multirow{2}{*}{1}&Stem layer: 16$\times$16 conv w/ s16,&\texttt{ConViT-B} \cite{convit}, \texttt{CrossViT-B} \cite{crossvit},\texttt{DeiT-B} \cite{deit}, \texttt{DeiT-S} \cite{deit},\\
&No Hierarchical, GeLU, LN, Final GAP&\texttt{ViT-S (patch size 16)} \cite{vit},\texttt{ResMLP-S24} \cite{resmlp}, \texttt{gMLP-S} \cite{gmlp}\\
\midrule
\multirow{3}{*}{2}&Stem layer: 4$\times$4 conv w/ s4, LN, GeLU, &\texttt{Twins-PCPVT-B} \cite{twins}, \texttt{Twins-SVT-S} \cite{twins}, \texttt{CoaT-Lite Small} \cite{coat},\\
&Transformer, No pooling at stem& \texttt{NesT-T} \cite{nest}, \texttt{Swin-T} \cite{swin}, \texttt{S3 (Swin-T)} \cite{s3}, \texttt{ConvNeXt-T} \cite{convnext},\texttt{ResMLP-B24} \cite{resmlp}\\
\midrule
\multirow{2}{*}{3}&Transformer, Final GAP, GeLU,&\texttt{XCiT-M24} \cite{xcit}, \texttt{XCiT-T12} \cite{xcit}, \texttt{HaloRegNetZ-B}\footnotemark[2], \texttt{TNT-S} \cite{tnt},\\
& Pooling at stem, InRes: 224 & \texttt{Visformer-S} \cite{visformer}, \texttt{PiT-S} \cite{pit}, \texttt{PiT-B} \cite{pit}\\
\midrule
\multirow{2}{*}{4}&Stem layer: stack of 3$\times$3 conv, 2D SA,&\texttt{HaloNet-50} \cite{halonet}, \texttt{LambdaResNet-50} \cite{lambdanet}, \texttt{BoTNeT-26} \cite{botnet},\\
& InRes: 256, Pooling at stem, SiLU&\texttt{GC-ResNeXt-50} \cite{gcnet}, \texttt{ECAHaloNeXt-50}\footnotemark[2], \texttt{ECA-BoTNeXt-26}\footnotemark[2]\\
\midrule
\multirow{2}{*}{5}&Stem layer: stack of 3$\times$3 convs, InRes: 256,&\texttt{LamHaloBoTNet-50}\footnotemark[2], \texttt{SE-BoTNet-33}\footnotemark[2], \texttt{SE-HaloNet-33}\footnotemark[2], \\
&2D SA, CWA: middle of blocks, CNN&\texttt{Halo2BoTNet-50}\footnotemark[2], \texttt{GC-ResNet-50} \cite{gcnet}, \texttt{ECA-Net-33} \cite{ecanet}\\
\midrule
\multirow{4}{*}{6}&Stem layer: 7$\times$7 conv w/ s2, ReLU, &\texttt{ResNet-50} \cite{resnet}, \texttt{ResNet-101} \cite{resnet}, \texttt{ResNeXt-50} \cite{resnext}, \\
&Pooling at stem, CNN, BN&\texttt{Wide ResNet-50} \cite{wideresnet}, \texttt{SE-ResNet-50} \cite{senet}, \texttt{SE-ResNeXt-50} \cite{senet},  \\
&&\texttt{ResNet-V2-50} \cite{resnetv2}, \texttt{ResNet-V2-101} \cite{resnetv2}, \texttt{ResNet-50 (GN)} \cite{wu2018groupnorm}, \\
&&\texttt{ResNet-50 (BlurPool)} \cite{blurpool}, \texttt{DPN-107} \cite{dpn}, \texttt{Xception-65} \cite{xception}\\
\midrule
\multirow{2}{*}{7}&NAS, Stem layer: 3$\times$3 conv w/ s2&\texttt{EfficientNet-B2} \cite{efficientnet}, \texttt{FBNetV3-G} \cite{fbnetv3}, \texttt{ReXNet} ($\times$1.5) \cite{rexnet}, \\
&CWA: middle of blocks, CWA, DW Conv &\texttt{RegNetY-32} \cite{regnetxy}, \texttt{MixNet-XL} \cite{mixnet}, \texttt{NF-RegNet-B1} \cite{nf-resnet-regent}\\
\midrule
\multirow{3}{*}{8}&Input resolution: 224, Stem layer: stack of 3$\times$3 convs,&\texttt{NFNet-L0}\footnotemark[2], \texttt{ECA-NFNet-L0}\footnotemark[2], \texttt{PoolFormer-M48} \cite{poolformer}, \\
&Group Conv, Final GAP, 2D SA&\texttt{ResNeSt-50}~\cite{resnest}, \texttt{ResNet-V2-50-D-EVOS}\footnotemark[2], \\
&&\texttt{ConvMixer-1536/20} \cite{convmixer}\\
\midrule
\multirow{3}{*}{9}&ReLU, Input resolution: 224, DW Conv, BN,&\texttt{ViT-B (patch size 32)} \cite{vit}, \texttt{R26+ViT-S} \cite{rvit}, \texttt{DLA-X-102} \cite{dla},\\
&2D self-attention&\texttt{eSE-VoVNet-39} \cite{esevovnet}, \texttt{ResNet-101-C} \cite{bagoftricks}, \texttt{RegNetX-320} \cite{regnetxy}, \\
&&\texttt{HRNet-W32} \cite{hrnet}\\
\midrule
\multirow{2}{*}{10}&ReLU + Leaky ReLU, InRes: 256, &\texttt{CSPResNet-50} \cite{cspnet}, \texttt{CSPResNeXt-50} \cite{cspnet}, \texttt{CSPDarkNet-53} \cite{yolov4}, \\
&Stem layer: 7$\times$7 conv, CNN, Pooling at stem&\texttt{NF-ResNet-50}~\cite{nf-resnet-regent}\\
\bottomrule
\end{tabular}
}
\end{table}

\renewcommand{\thefootnote}{\fnsymbol{footnote}}
\footnotetext[2]{\scriptsize Customized models by \cite{rw2019timm}: \texttt{HaloRegNetZ} = \texttt{HaloNet} + \texttt{RegNetZ}; \texttt{ECA-BoTNeXt} = \texttt{ECA-Net} + \texttt{HaloNet} + \texttt{ResNeXt}; \texttt{ECA-BoTNeXt} = \texttt{ECA-Net} + \texttt{BoTNet} + \texttt{ResNeXt}; \texttt{LamHaloBoTNet} = \texttt{LambdaNet} + \texttt{HaloNet} + \texttt{BoTNet}; \texttt{SE-BoTNet} = \texttt{SENet} + \texttt{BoTNet}; \texttt{SE-HaloNet} = \texttt{SENet} + \texttt{HaloNet}; \texttt{Halo2BoTNet} = \texttt{HaloNet} + \texttt{BoTNet}; \texttt{NFNet-L0} = an efficient variant of \texttt{NFNet-F0} \cite{nfnet}; \texttt{ECA-NFNet-L0} = \texttt{ECA-Net} + \texttt{NFNet-L0}; %
\texttt{ResNet-V2-D-EVOS} = \texttt{ResNet-V2} + \texttt{EvoNorms} \cite{evos}.}%
\renewcommand{\thefootnote}{\arabic{footnote}}

\paragraph{Clustering analysis.}
We additionally provide a clustering analysis based on the architectural similarities. We construct a pairwise similarity graph with adjacency matrix $A$ between all 69 architectures where its vertex denotes an architecture, and its edge denotes the similarity between two networks. We perform the spectral clustering \cite{ng2001spectral} on $A$ where the number of clusters $K$ is set to 10: We compute the Laplacian matrix of $A$,
$L = D-A$ where $D$ is the diagonal matrix and its $i$-th component is $\sum_j A_{ij}$. Then, we perform K-means clustering on the $K$-largest eigenvectors of $L$.
The pairwise distances of spectral features (\ie, 10-largest eigenvectors of $L$) of 69 neural architectures are shown in \cref{fig:similarity-different-vs-same}. The rows and columns of \cref{fig:similarity-different-vs-same} are sorted by the clustering index (\cref{tab:clustering-results}). More details with model names are described in \cref{subsec:appendix-similarities}. We can see the block-diagonal patterns, \ie, in-clusters similarities are more significant than between-clusters similarities. More details are in \cref{subsec:appendix-spectral-clustering}.

\cref{tab:clustering-results} shows the clustering results on 69 networks and the top-5 keywords for each cluster based on term frequency-inverse document frequency (TF-IDF) analysis. Specifically, we treat each model feature as a word and compute TF and IDF by treating each architecture as a document. Then we compute the average TF-IDF for each cluster and report top-5 keywords. Similar to \cref{fig:model-analysis}, the base architecture (\eg, CNN in Cluster 5, 6, 10 and Transformer in Cluster 2, 3) and the design choice for the stem layer (\eg, Cluster 1, 2, 4, 5, 6, 7, 8, 10) repeatedly appear at the top keywords. Especially, we can observe that the differences in base architecture significantly cause the diversity in model similarities, \eg, non-hierarchical Transformers (Cluster 1), hierarchical networks with the patchification stem (Cluster 2), hierarchical Transformers (Cluster 3), CNNs with 2D self-attention (Cluster 4, 5), ResNet-based architectures (Cluster 6), and NAS-based architectures (Cluster 7).

\begin{figure}[t]
\begin{minipage}[t]{0.43\linewidth}
\captionof{table}{\textbf{\ours within the same architecture.} We compare the average similarity within the same architecture but trained with different procedures,
``All'' denotes the average similarity of 69 architectures.}
\label{tab:inner-similarity}
\centering
\small
\begin{tabular}{c|cc}
\toprule
Architecture&\texttt{ResNet-50}&\texttt{ViT-S}\\
\midrule
Init&4.23&4.21\\
Hparam&4.05&4.22\\
Tr. Reg.&3.27&3.44\\
\midrule
{All}&\multicolumn{2}{c}{2.73}
\\
\bottomrule
\end{tabular}
\end{minipage}\hfill
\begin{minipage}[t]{.55\linewidth}
\centering
\captionof{table}{\textbf{Ensemble performance with diverse architectures.} We report the error reduction rate by varying the number of ensembled models and the diversity of the ensemble models (related to \cref{fig:ens_perf_a}). ``rand'' indicates the random choice of models. }
\label{tab:ens1}
\begin{tabular}{cc|ccccc|c}
& & \multicolumn{5}{c|}{\tiny less diverse $\leftarrow$ \# of clusters $\rightarrow$ more diverse}&\\
& & 1 & 2 & 3 & 4 & 5&rand \\
\midrule
\parbox[t]{2mm}{\multirow{4}{*}{\rotatebox[origin=c]{90}{{\scriptsize \# of models}}}} &2 & 7.13 & \textbf{7.84} & & & & 7.78\\
&3 & 10.17 & 10.84 & \textbf{11.20} & & & 11.11\\
&4 & 11.70 & 12.45 & 12.80 & \textbf{13.00} & 
 & 12.90\\
&5 & 12.58 & 13.41 & 13.79 & 13.99 & \textbf{14.11} & 14.01\\
\end{tabular}
\end{minipage}
\end{figure}

\subsection{The Relationship between Training Strategy and SAT}
\label{subsec:trainging-strategy-similarity}
The architectural difference is not the only cause of the model diversity. We compare the impact by different architecture choices (\eg, ResNet and ViT) and by different training strategies while fixing the model architecture, as follows: \textbf{Different initializations} can affect the model training by the nature of the stochasticity of the training procedure. For example, \citet{somepalli2022reproducibility} showed that the decision boundary of each architecture could vary by different initializations. We also consider \textbf{different optimization hyper-parameters} (\eg, learning rate, weight decay). Finally, we study the effect of \textbf{different training regimes} (\eg, augmentations, type of supervision). 
For example, the choice of data augmentation \cite{zhang2017mixup,yun2019cutmix} or label smoothing \cite{szegedy2016rethinking}
can theoretically or empirically affect adversarial robustness \cite{zhang2020does, park2022msda, shafahi2019label, chun2019icmlw}.
We also investigate the effect of supervision, such as self-supervision \cite{mocov3,mae,byol} or semi-weakly supervised learning \cite{yalniz2019billion}.
Note that the training strategies inevitably contain the former ones.
For example, when we train models with different training regimes, models have different initialization seeds and different optimization hyper-parameters. Comparing 69 different architectures also contains the effect of different initialization and optimization hyper-parameters and parts of different training regimes. This is necessary for achieving high classification performance.

\cref{tab:inner-similarity} shows the comparison of similarity scores between the same architecture but different learning methods (a smaller similarity means more diversity).
We report two architectures, ResNet-50 and ViT-S, and their training settings are in \cref{sec:appendix-different-training-settings}. We also show the average SAT between all 69 architectures. In the table, we first observe that using different random initialization or different optimization hyper-parameters shows high correlations with each other (almost $\geq$\,4.2) while the average similarity score between various neural architectures is 2.73. In other words, the difference in initializations or optimization hyper-parameters does not significantly contribute to the model diversity. 

\begin{figure}[t]
\begin{minipage}[b]{0.55\linewidth}
    \centering
    \includegraphics[width=0.6\linewidth]{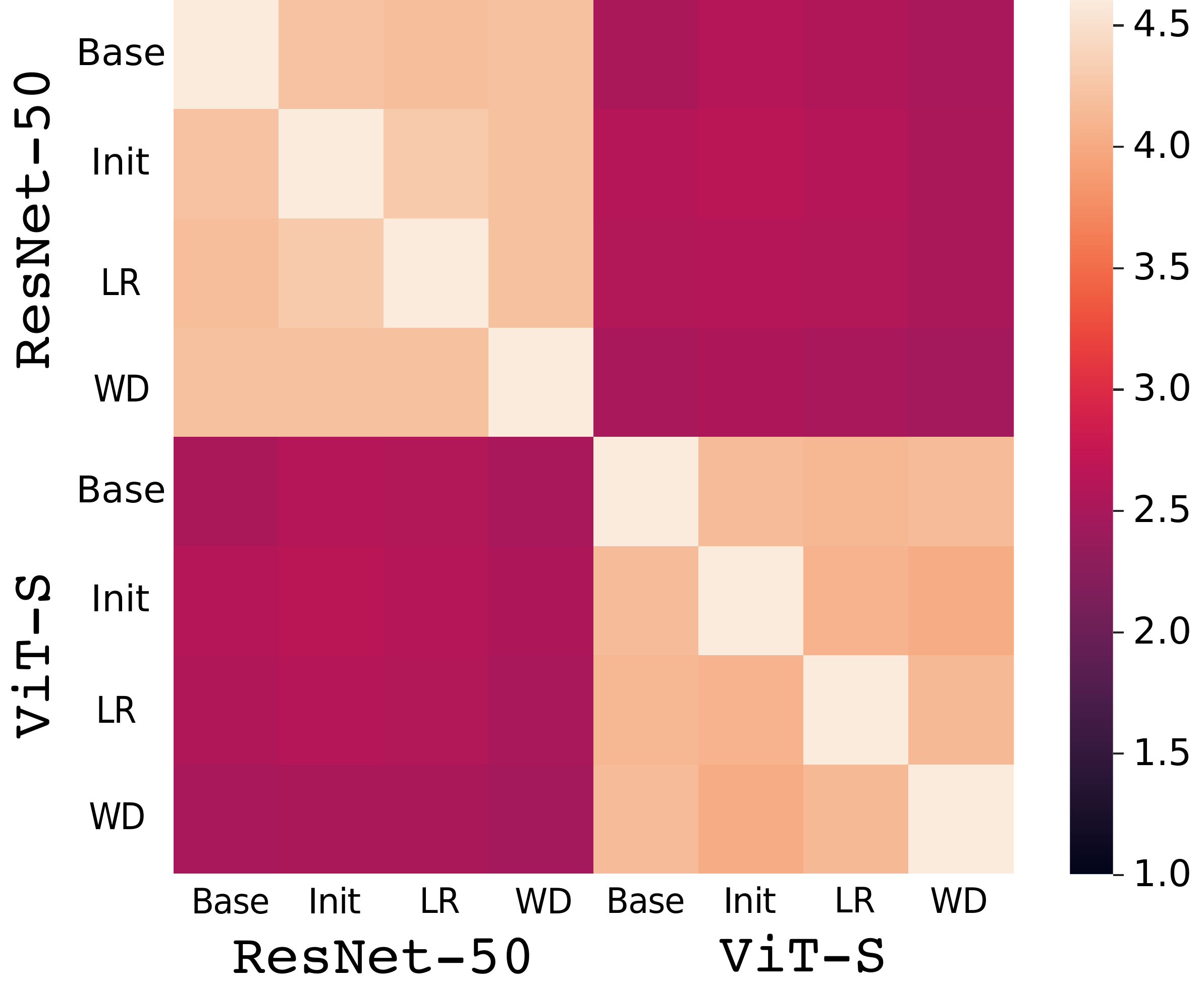}
    {\caption{\small {\bf Pairwise distance of spectral features by different optimizations.} \texttt{Init}, \texttt{LR}, and \texttt{WD} are randomly chosen from models trained with different settings of initialization, learning rate, and weight decay in \cref{tab:inner-similarity}.}
    \label{fig:similarity-different-vs-same-b}}
\end{minipage}\hspace{0.03\textwidth}
\begin{minipage}[b]{0.35\textwidth}
    \centering
    \includegraphics[width=\linewidth]{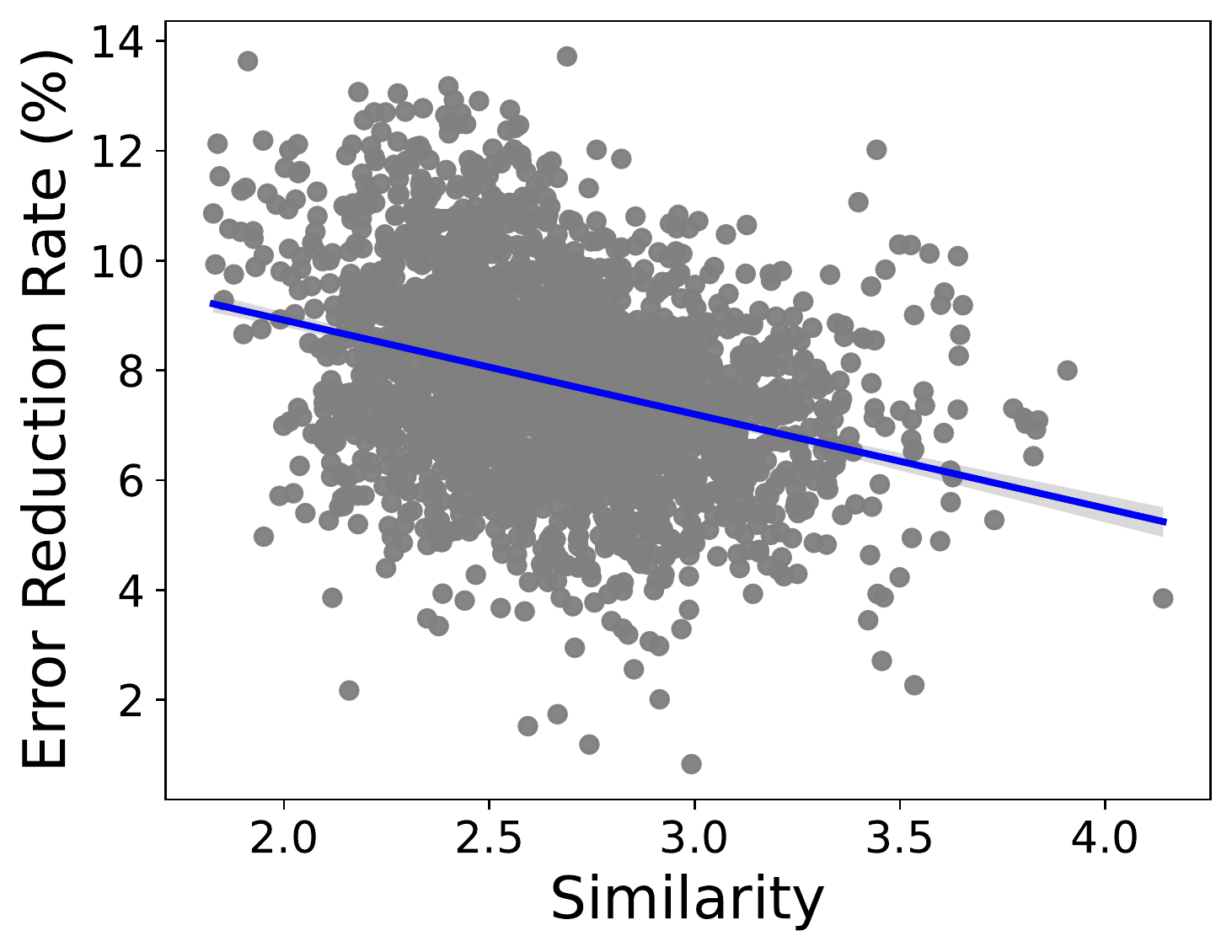}%
    \caption{\small {\bf Correlation between pairwise \ours and ensemble performance.} The trend line and its 90\% confidence interval are shown.}
    \label{fig:trans-ei}
\end{minipage}
\end{figure}

Second, we observe that using different learning techniques
remarkably affects SAT (3.27 for ResNet and 3.44 for ViT), but is not as significant as the architectural difference (2.73).
Furthermore, the change of SAT caused by different initializations or hyper-parameters is less marked than the change caused by different architecture (\cref{fig:similarity-different-vs-same-b}). These observations provide two insights.
First, %
the diversity resulting from various training strategies
is not significant enough compared to the diversity of architecture.
Second, designing new architecture is more efficient in achieving diverse models rather than re-training the same one.

\section{SAT Applications with Multiple Models}
\label{sec:applications}

Here, we analyze how SAT is related to downstream tasks involving more than one model. First, we show that using more diverse models will lead to better ensemble performance. Second, we study the relationship between knowledge distillation and \ours. Furthermore, we can suggest a similarity-based guideline for choosing a teacher model when distilling to a specific architecture. Through these observations, we can provide insights into the necessity of diverse models.

\subsection{Model Diversity and Ensemble}
\label{subsec:ensemble}

\paragraph{Settings.}
The model ensemble is a practical technique for achieving high performance. However, only few works have studied the relationship between ensemble performance and model similarity, particularly for large-scale complex models. 
Previous studies are mainly conducted on tiny datasets and linear models \cite{kuncheva2003measures}. 
We investigate the change of ensemble performance by the change of similarity based on the unweighted average method \cite{ju2018relative} (\ie, averaging the logit values of the ensembled models). Because the ensemble performance is sensitive to the original model performances, we define Error Reduction Rate (ERR) as $1-\frac{\text{Err}_\text{ens}(M)}{\frac{1}{|M|} \sum_{m \in M}{\text{Err}(m)}}$, where $M$ is the set of the ensembled models, ERR($m$) denotes the top-1 ImageNet validation error of model $m$, and Err$_\text{ens}$($\cdot$) denotes the top-1 error of the model ensemble results.

\paragraph{Results.}
We first measure the 2-ensemble performances among the 69 architectures (\ie, the number of ensembles is $\binom{69}{2}=2346$). We plot the relationship between SAT and ERR in \cref{fig:trans-ei}. We observe that there exists a strong negative correlation between the model similarity and the ensemble performance (Pearson correlation coefficient $-0.32$ with p-value $\approx 0$
and Spearman correlation $-0.32$ with p-value $\approx 0$,
\ie, \emph{more diversity leads to better ensemble performance}.

\begin{figure}[t]
	\centering
    \begin{subfigure}[b]{0.33\linewidth}
        \includegraphics[width=\linewidth]{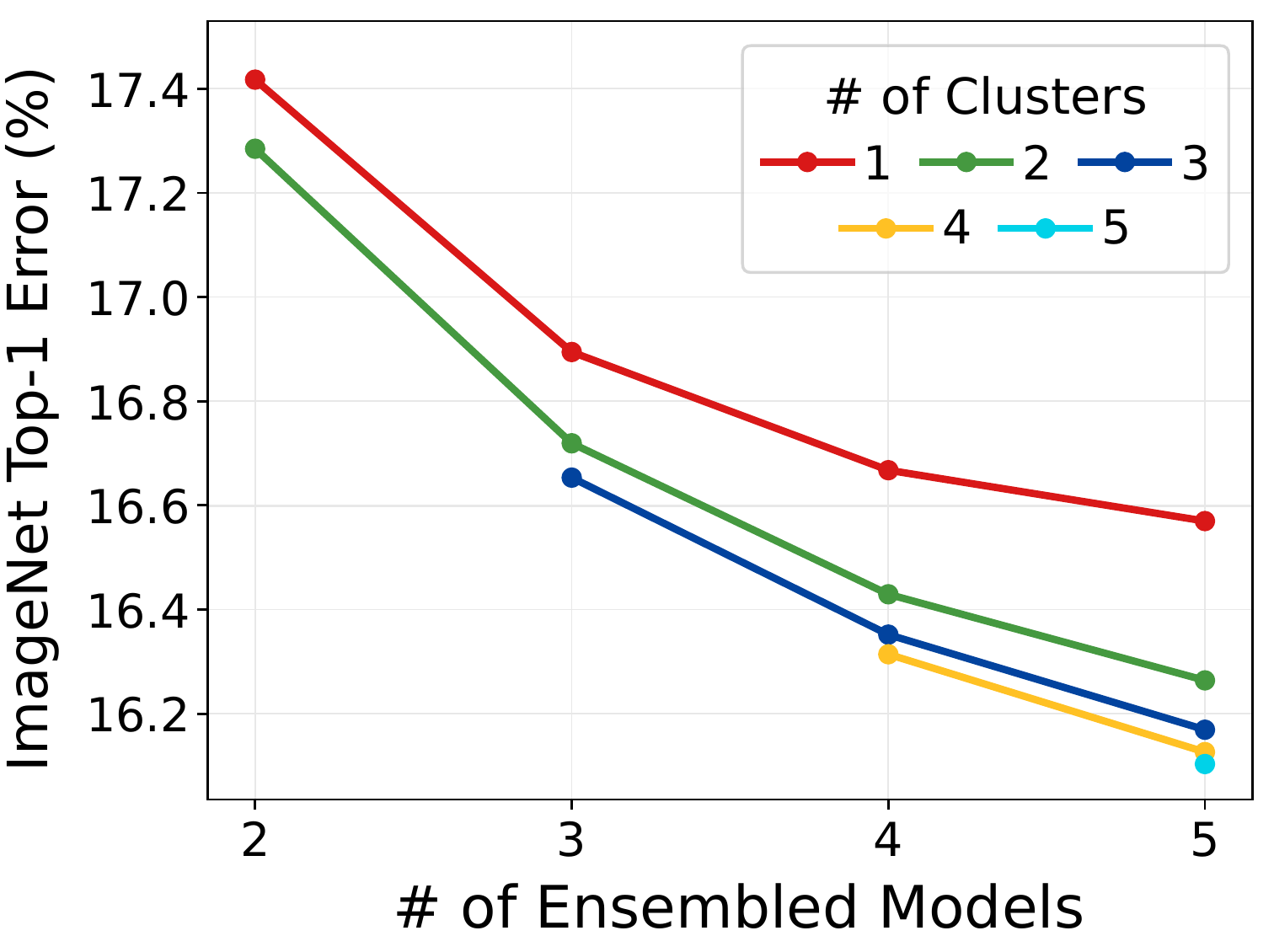}
        \caption{\small Top-1 Error}
        \label{fig:ens_perf_a}
	\end{subfigure}%
	\begin{subfigure}[b]{0.33\linewidth}
        \includegraphics[width=\linewidth]{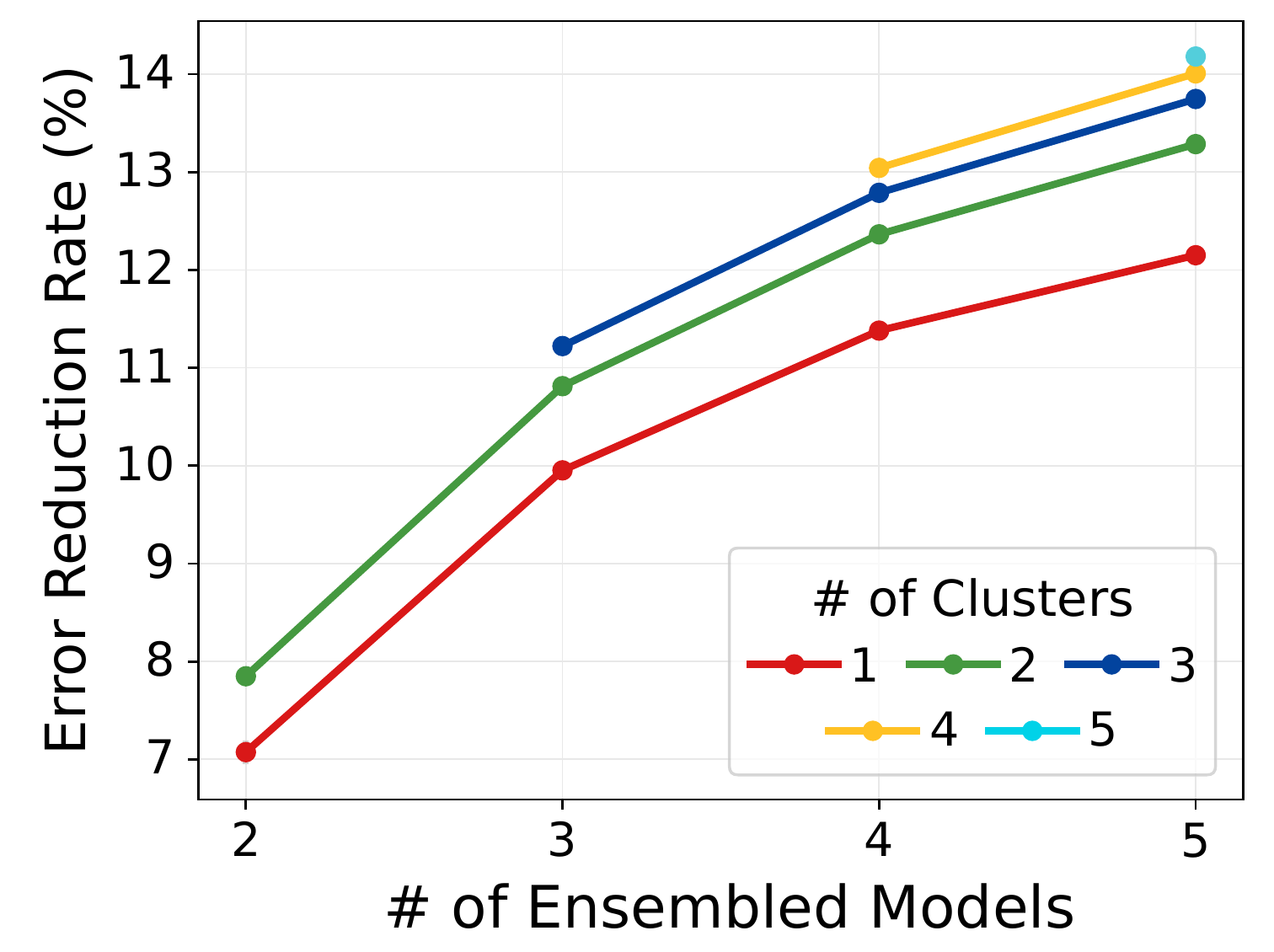}
        \caption{\small Error reduction rate}
        \label{fig:ens_perf_b}
	\end{subfigure}%
	\begin{subfigure}[b]{0.33\linewidth}
        \includegraphics[width=\linewidth]{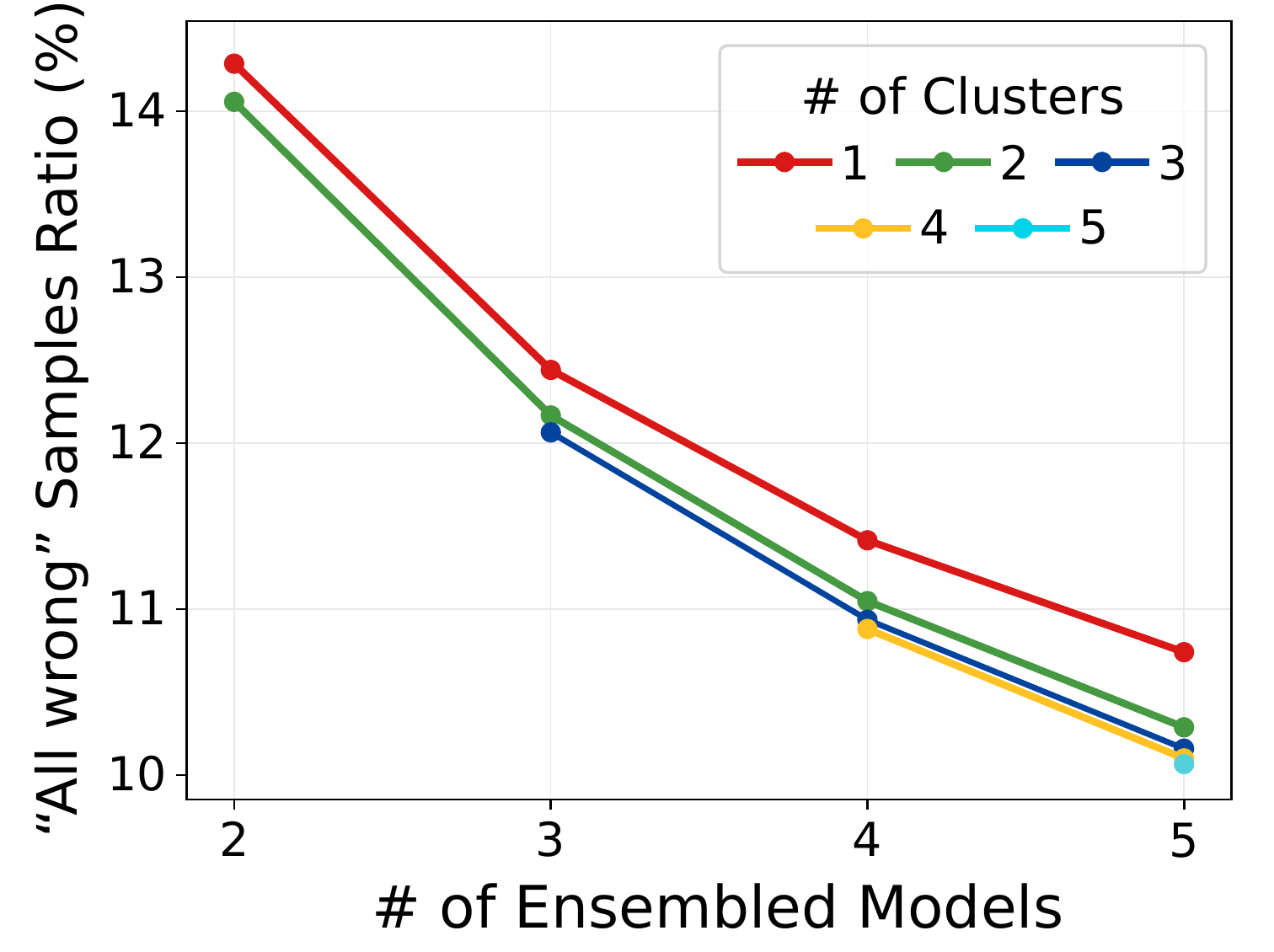}
        \caption{\small ``All wrong'' samples ratio}
        \label{fig:ens_perf_c}
	\end{subfigure}
	\caption{\small {\bf Model diversity
 and ensemble performance.} We report ensemble performances by varying the number of ensembled models ($N$) and the diversity of the models. The diversity is controlled by choosing the models from $k$ different clusters.}
	\label{fig:ensemble-performance}
\end{figure}

We also conduct $N$-ensemble experiments with $N\geq2$ based on our clustering results in \cref{tab:clustering-results}. We evaluate the average ERR of the ensemble of models from $k$ clusters, \ie, if $N=5$ and $k=3$, the ensembled models are only sampled from the selected 3 clusters while ignoring the other 7 clusters. We investigate the effect of model diversity and ensemble performance by examining $k=1 \ldots N$ (\ie, larger $k$ denotes more diverse ensembled models). We report the result with ImageNet top-1 error and ERR in \cref{fig:ens_perf_a} and \cref{fig:ens_perf_b}.

\begin{figure}[t]
\centering
    \begin{subfigure}[b]{.35\linewidth}
        \includegraphics[width=\linewidth]{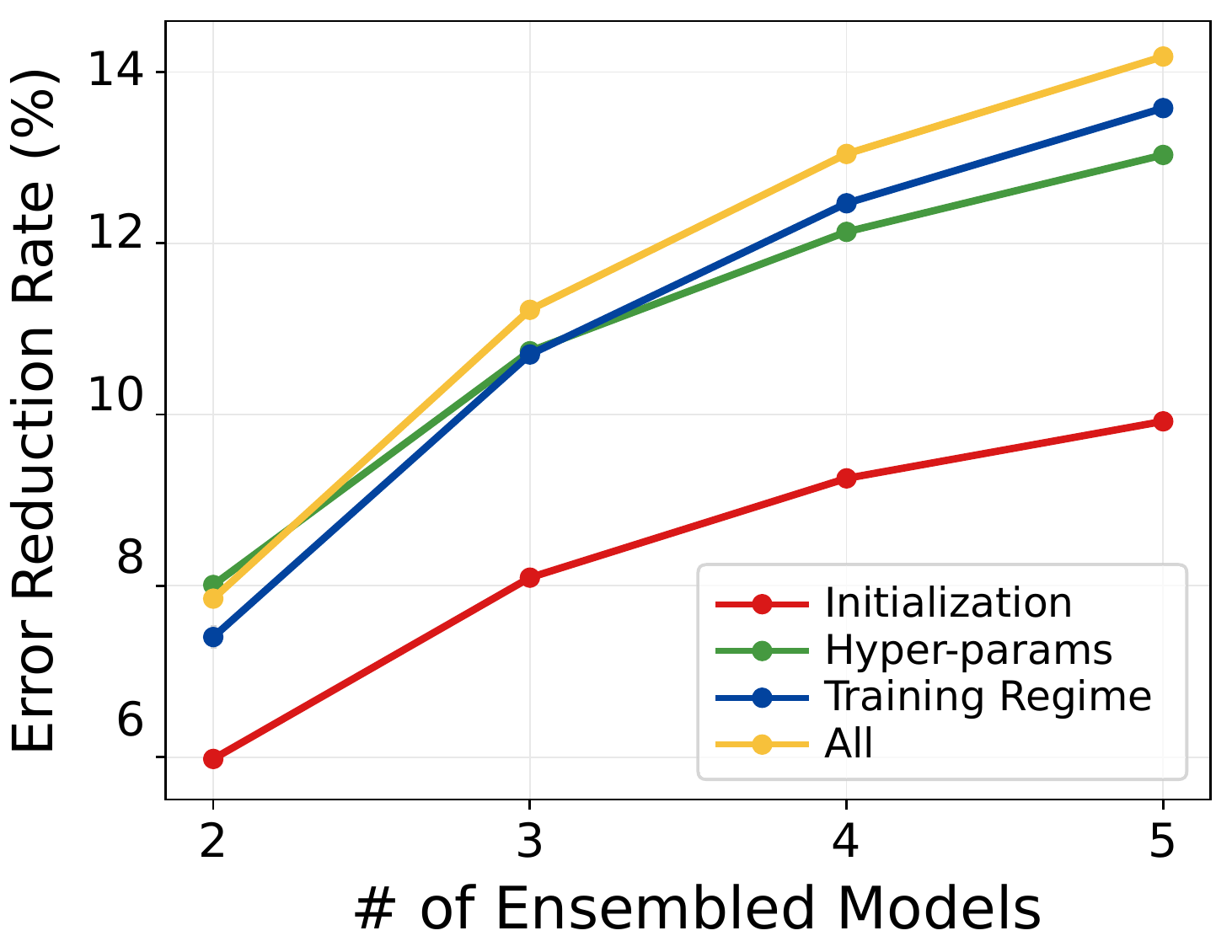}
        \caption{\small \texttt{ResNet-50}}
        \label{fig:ens_perf_resnetdeit-a}
	\end{subfigure} \hspace{1em}
	\begin{subfigure}[b]{.35\linewidth}
        \includegraphics[width=\linewidth]{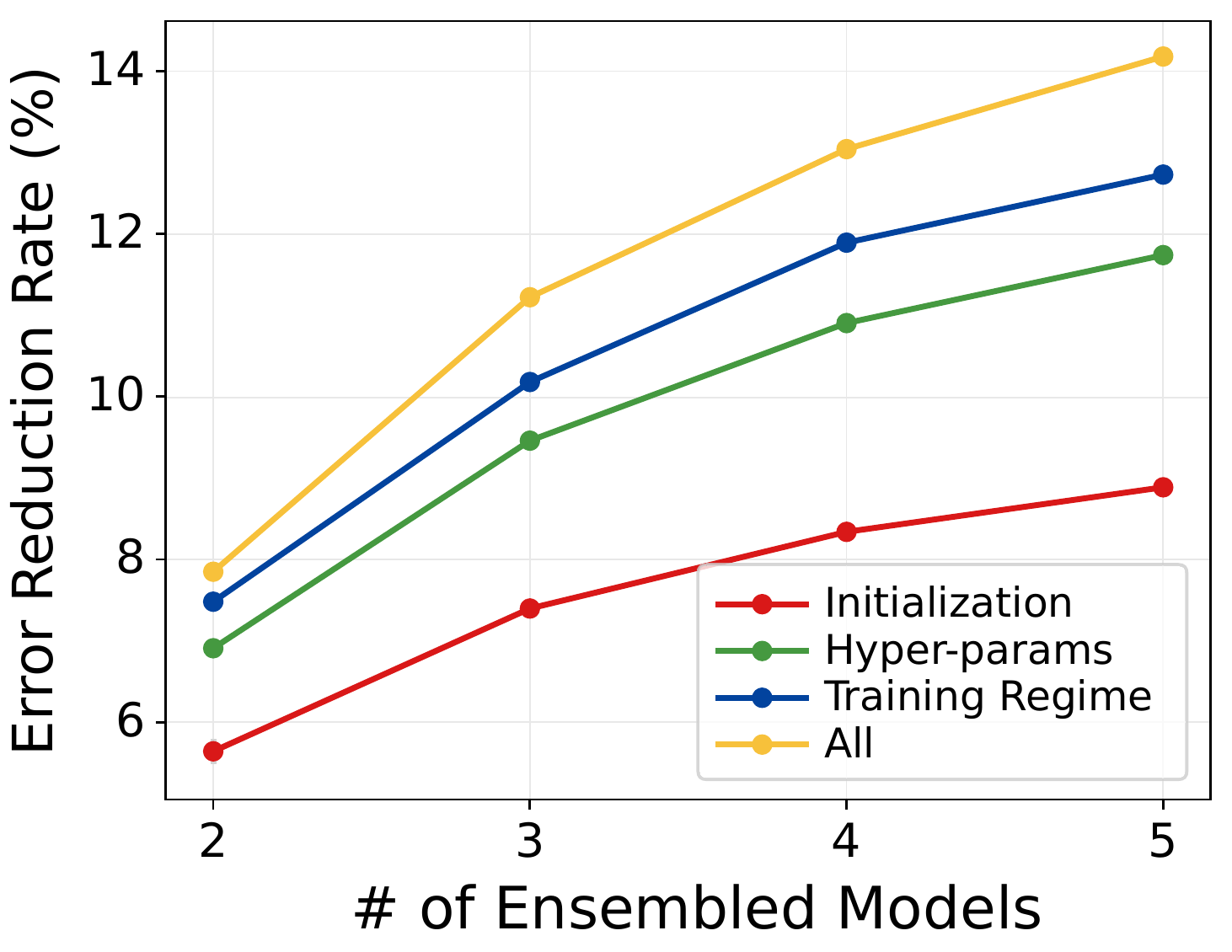}
        \caption{\small \texttt{ViT-S}}
        \label{fig:ens_perf_resnetdeit-b}
	\end{subfigure}
\caption{\small {\bf Diversity by training techniques and ensemble.} We report the the same metrics as \cref{fig:ensemble-performance} for various \texttt{ResNet-50} and \texttt{ViT-S} models in \cref{tab:inner-similarity}.}
\label{fig:ens_perf_resnetdeit}
\end{figure}
In all metrics, we observe that the ensemble of more diverse models shows better performance. Interestingly, \cref{fig:ens_perf_b} shows that when the number of clusters for the model selection ($k$) is decreased, the ensemble performance by the number of ensembled models ($N$)
quickly reaches saturation.
\cref{tab:ens1} shows that the ensemble performances by choosing the most diverse models via SAT always outperform the random ensemble.
Similarly, \cref{fig:ens_perf_c} shows that the number of wrong samples by all models is decreased by selecting more diverse models.

\paragraph{Training Strategy vs. Architecture in the ensemble scenario?}
Remark that \cref{tab:inner-similarity} showed that the different training strategies are not as effective as different architectures for diversity. To examine this on the ensemble scenario, we report the ensemble results of different training strategies, \ie, the same \texttt{ResNet-50} and \texttt{ViT-S} in \cref{tab:inner-similarity}. For comparison with different architectures, we also report the ensemble of different architectures where all ensembled models are from different clusters (\ie, $N$=$k$ in \cref{fig:ensemble-performance}). \cref{fig:ens_perf_resnetdeit} shows that although using diverse training regimes (blue lines) improves ensemble performance compared to other techniques (red and green lines), the improvements by using different architectures (yellow lines) are more significant than the improvements by using different training regimes (blue lines) with large gaps.

\begin{figure}[t]
\centering
    \begin{subfigure}[b]{0.48\linewidth}
        \includegraphics[width=\linewidth]{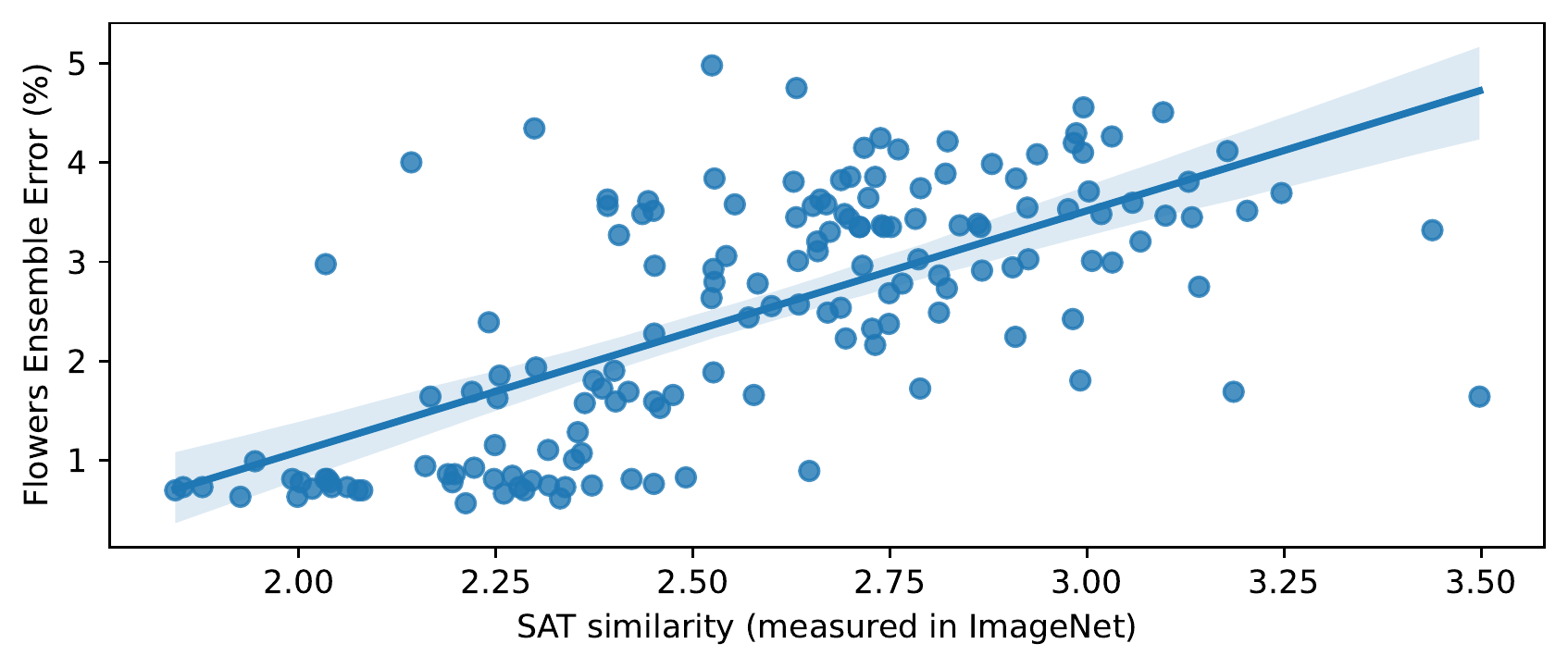}
        \caption{SAT vs. Flowers ensemble}
        \label{fig:ens_perf_flower}
	\end{subfigure}
	\begin{subfigure}[b]{0.48\linewidth}
        \includegraphics[width=\linewidth]{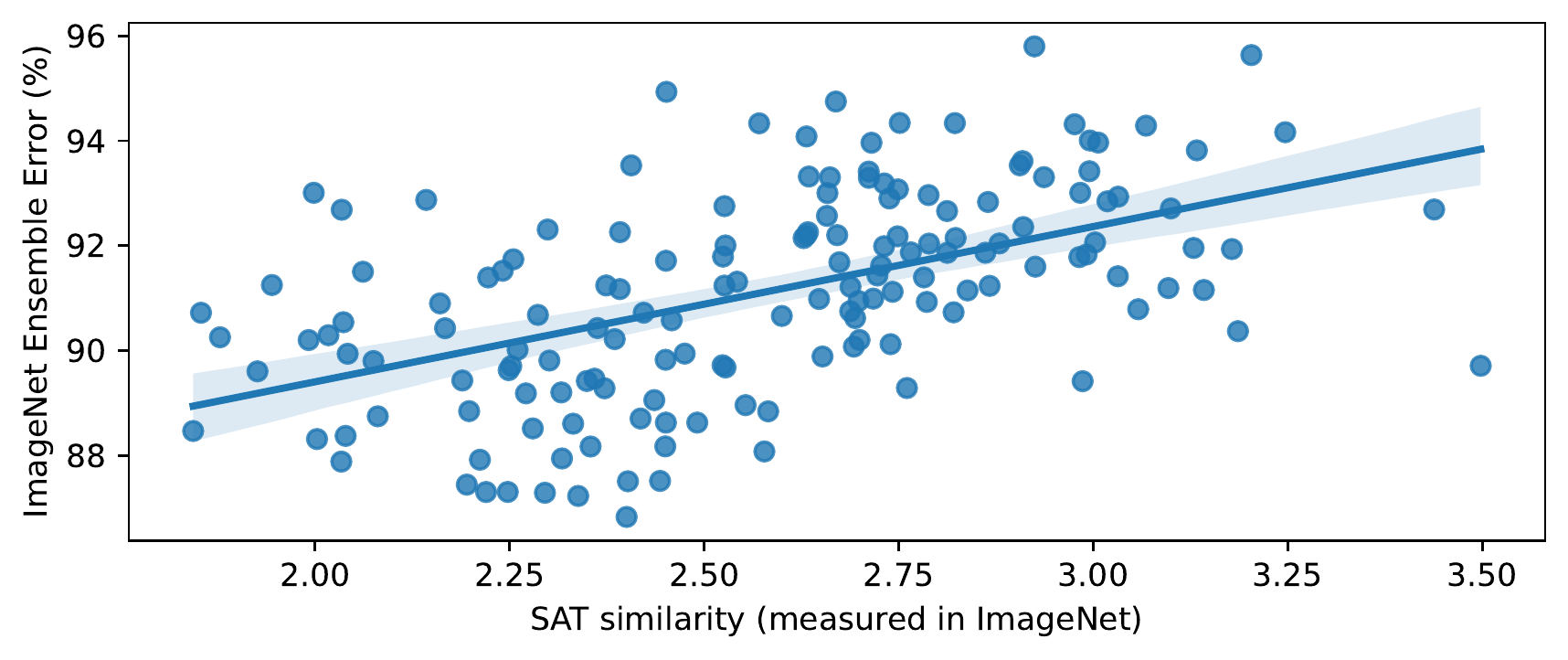}
        \caption{SAT vs. ImageNet ensemble}
        \label{fig:ens_perf_imgnet}
	\end{subfigure}
\caption{\small {\bf Cross-dataset SAT results.} SAT measured on ImageNet also has a positive correlation with ensemble performances on Flowers-102 \cite{flowers}.}
\label{fig:ens_perf_diff_data}
\end{figure}

\begin{figure}[t]
    \centering
  \includegraphics[width=\linewidth]{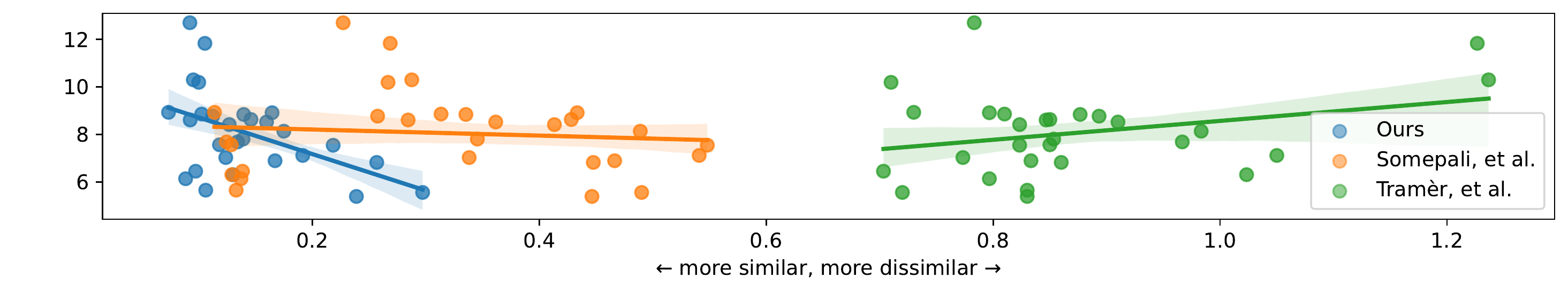}
    \caption{\small {\bf Empircial comparison.} Relationship between the model similarity, including \ours, \citet{somepalli2022reproducibility} and \citet{tramer2017space}, and 2-ensemble performance.}
    \label{fig:twoenscomp}        
\end{figure}

\paragraph{Generalizability to other datasets.}
We examine whether more diverse architectures in ImageNet SAT also lead to better ensemble performances on the other datasets. We fine-tuned all 69 architectures to the Flowers-102 dataset \cite{flowers}, and filter out low performing models (< 95\% top-1 accuracy). After the filtering, we have 16 fine-tuned models.
Using the fine-tuned models, we plot the relationship between the Flowers-102 ensemble performance and SAT score measured in ImageNet. \cref{fig:ens_perf_diff_data} shows that \ours also highly correlates with Flowers ensemble performances, despite that SAT is measured on ImageNet. This experimental result supports that SAT similarity can be applied in a cross-domain manner.

\paragraph{Comparison of different similarity functions in the ensemble scenario.} Finally, we compare the impact of the choice of the similarity function and the ensemble performance when following our setting. We compare \ours with \citet{somepalli2022reproducibility} and \citet{tramer2017space} on the 2-ensemble scenario with 8 out of 69 models due to the stability issue of \citet{tramer2017space}. \cref{fig:twoenscomp} shows the relationship between various similarity functions and the 2-ensemble performance. We observe that \ours only shows a strong positive correlation (blue line), while the others show an almost random or slightly negative correlation.
Finally, in \cref{subsec:appendix_empricial_comparison_ensemble}, we compare \ours with \citet{somepalli2022reproducibility} and a naive architecture-based clustering using our features under the same setting of \cref{fig:ensemble-performance} and \ref{fig:ens_perf_resnetdeit}. Similarly, \ours shows the best ensemble performance against the comparison methods.

\begin{figure}[b]
    \centering
    \begin{subfigure}[b]{0.32\linewidth}
    \includegraphics[width=\linewidth]{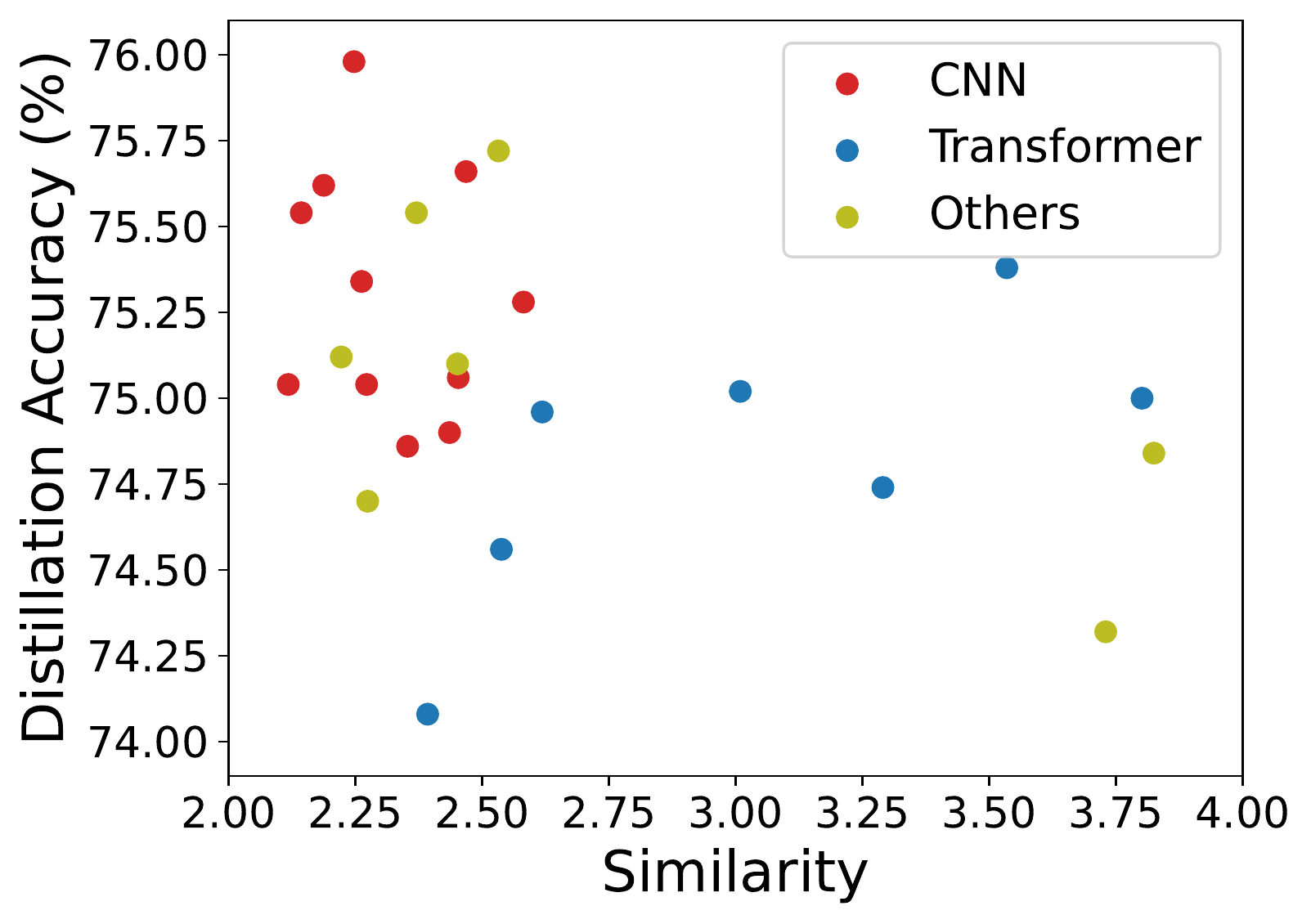}
    \caption{Results on various teachers}
    \label{fig:deit-distill-a}
    \end{subfigure}
    \begin{subfigure}[b]{0.32\linewidth}
    \includegraphics[width=\linewidth]{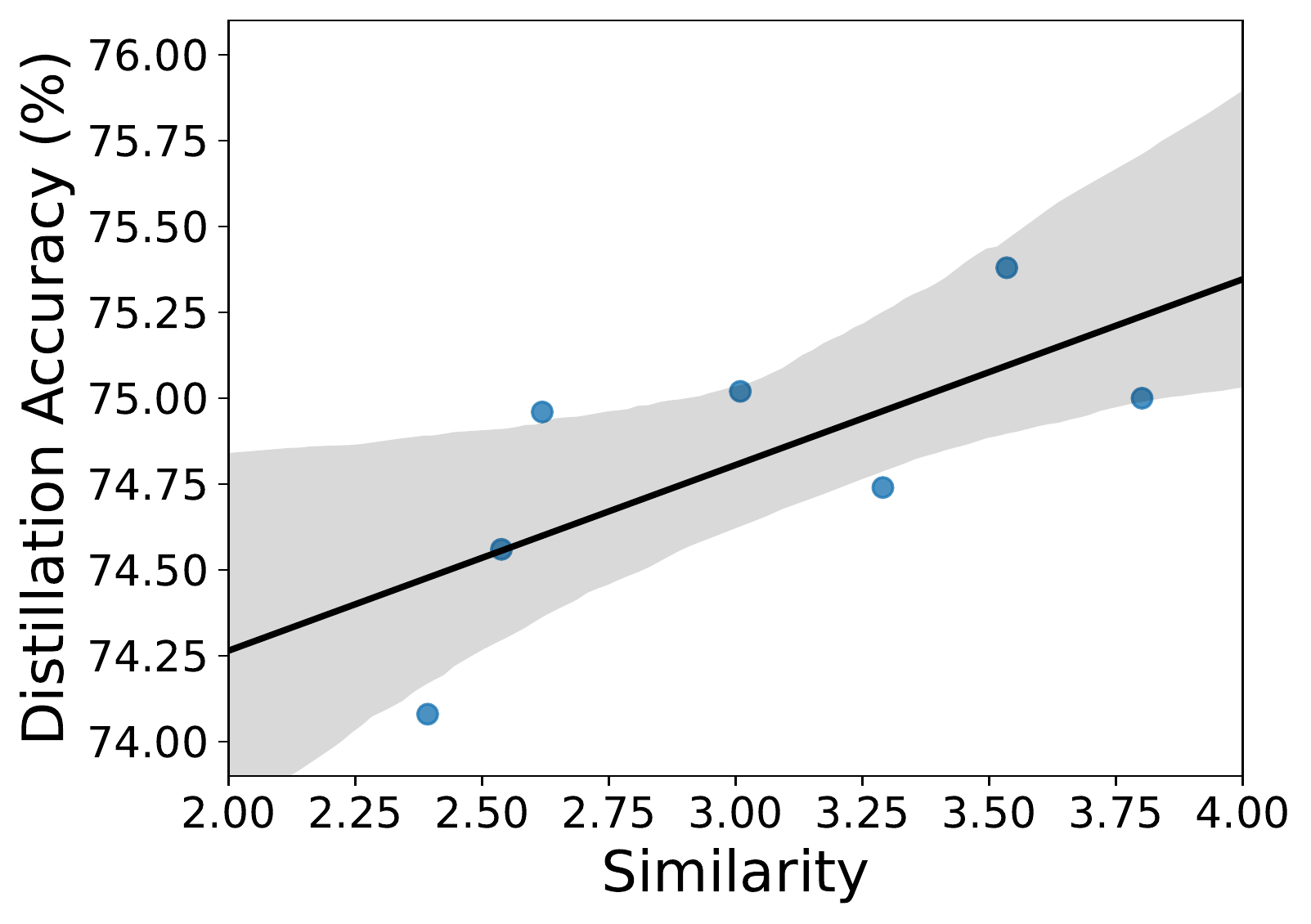}
    \caption{Transformer teachers}
    \label{fig:deit-distill-b}
    \end{subfigure}
    \begin{subfigure}[b]{0.32\linewidth}
    \includegraphics[width=\linewidth]{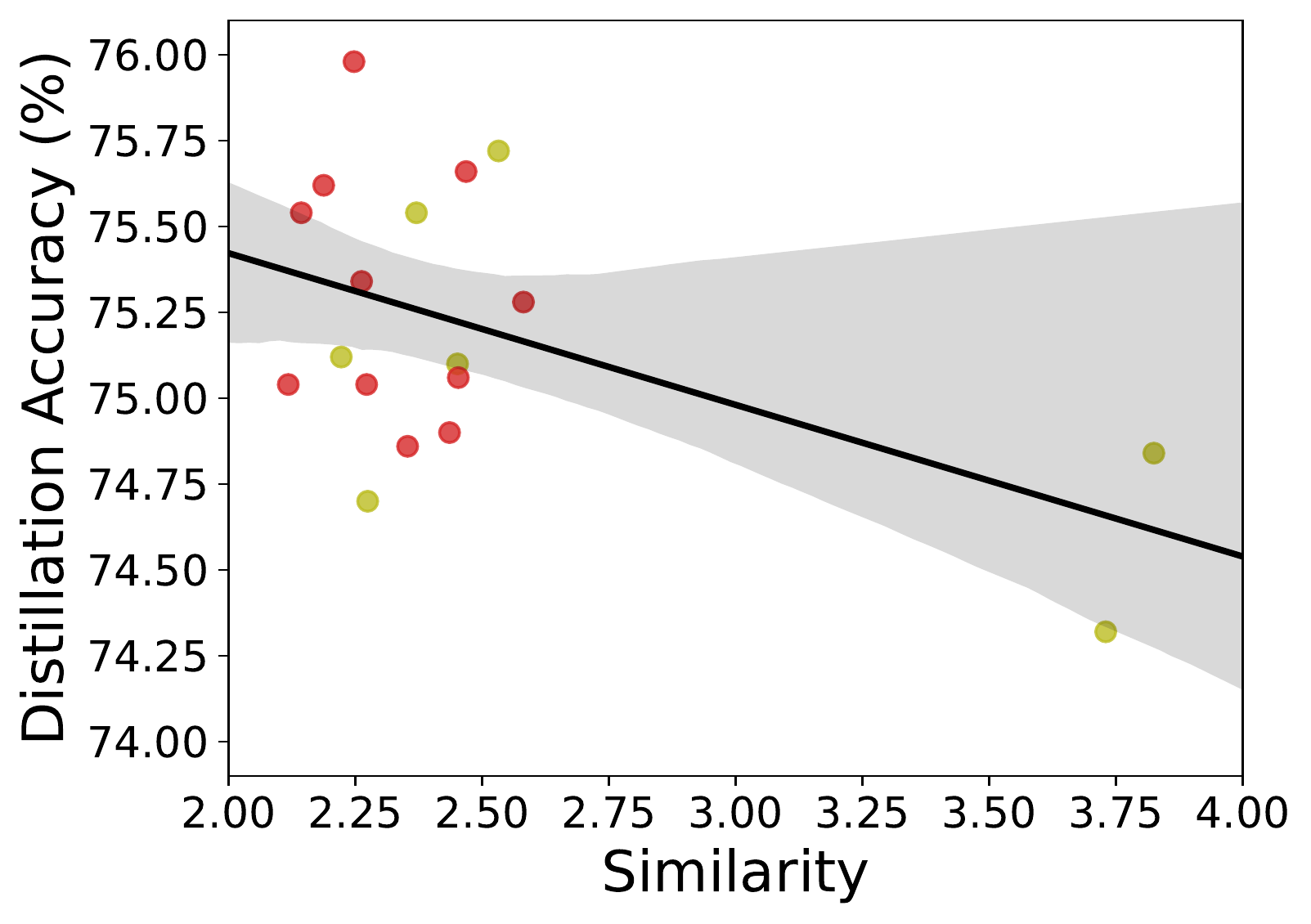}
    \caption{CNN \& other teachers}
    \label{fig:deit-distill-c}
    \end{subfigure}
    \caption{\small {\bf Model diversity and distillation performance.} (a) We show the relationship between teacher-student similarity and distillation performance of 25 \texttt{DeiT-S} models distilled by various teacher networks.
    We show the relationship when the teacher and student networks are based on (b) Transformer and (c) otherwise.}
    \label{fig:deit-distill}
\end{figure}

\subsection{Model Diversity and Knowledge Distillation}
\label{subsec:distill}

Knowledge distillation (KD) \cite{hinton2015distilling} is a training method for transferring rich knowledge of a well-trained teacher network. Intuitively, KD performance affects a lot by choice of the teacher network; however, the relationship between similarity and KD performance has not yet been explored enough, especially for ViT. This subsection investigates how the similarity between teacher and student networks contributes to the distillation performance. There are several studies showing two contradictory conclusions; Jin \etal \cite{jin2019knowledge} and Mirzadeh \etal \cite{mirzadeh2020improved} showed that a similar teacher leads to better KD performance; Touvron \etal \cite{deit} reports that distillation from a substantially different teacher is beneficial for ViT.

We train 25 \texttt{ViT-Ti} models with different teacher networks from 69 models that we used by the hard distillation strategy \cite{hinton2015distilling}. Experimental details are described in Appendix.
\cref{fig:deit-distill-a} illustrates the relationship between the teacher-student similarity and the distillation performance. 
\cref{fig:deit-distill-a} tends to show a not significant negative correlation between teacher-student similarity and distillation performance ($-$0.32 Pearson correlation coefficient with 0.12 p-value). However, if we only focus on when the teacher and student networks are based on the same architecture (\ie, Transformer), we can observe a strong positive correlation (\cref{fig:deit-distill-b}) -- 0.70 Pearson correlation coefficient with 0.078 p-value. In this case, our observation is aligned with \cite{jin2019knowledge,mirzadeh2020improved}: a teacher similar to the student improves distillation performance. However, when the teacher and student networks are based on different architectures (\eg, CNN), then we can observe a stronger negative correlation (\cref{fig:deit-distill-c}) with $-$0.51 Pearson correlation coefficient and 0.030 p-value. In this case, a more dissimilar teacher leads to better distillation performance. We also test other factors that can affect distillation performance in Appendix; We observe that distillation performance is not correlated to teacher accuracy in our experiments.

Why do we observe contradictory results for Transformer teachers (\cref{fig:deit-distill-b}) and other teachers (\cref{fig:deit-distill-c})? Here, we conjecture that when the teacher and student networks differ significantly, distillation works as a strong regularizer. In this case, using a more dissimilar teacher can be considered a stronger regularizer (\cref{fig:deit-distill-c}). On the other hand, we conjecture that if two networks are similar, then distillation works as easy-to-follow supervision for the student network. In this case, a more similar teacher will work better because a more similar teacher will provide more easy-to-follow supervision for the student network (\cref{fig:deit-distill-b}). Our experiments show that the regularization effect improves distillation performance better than easy-to-follow supervision (\ie, the best-performing distillation result is by a CNN teacher). Therefore, in practice, we recommend using a significantly different teacher network for achieving better distillation performance (\eg, using RegNet \cite{regnetxy} teacher for ViT student as \cite{deit}).

\section{Discussion}
In \cref{sec:appendix-limitations-discussions}, we describe more discussions related to \ours. We first propose an efficient approximation of \ours when we have a new model; instead of generating adversarial samples from all models, only generating adversarial samples from the new model can an efficient approximation of \ours (\cref{subsec:efficient_approx}). We also show that \ours and the same misclassified samples have a positive correlation in \cref{subsec:aat_and_misclsf}. \cref{subsec:appendix_sat_during_training} demonstrates that we can estimate the similarity with a not fully trained model (\eg, a model in an early stage). Finally, we describe more possible applications requiring diverse models (\cref{subsec:appendix_more_apps}).

\section{Conclusion}
We have explored similarities between image classification models to investigate what makes the model similar or diverse and whether developing and using diverse models is required. For quantitative and model-agnostic similarity assessment, we have suggested a new similarity function, named \ours, based on attack transferability demonstrating differences in input gradients and decision boundaries.
Using \ours, we conduct a large-scale and extensive analysis using 69 state-of-the-art ImageNet models.
We have shown that macroscopic architectural properties, such as base architecture and stem architecture, have a more significant impact on similarity than microscopic operations, such as types of convolution, with numerical analysis.
Finally, we have provided insight into the ML applications using multiple models based on SAT, \eg, model ensemble or knowledge distillation.
Overall, we suggest using SAT to improve methods with multiple models in a practical scenario with a large-scale training dataset and a highly complex architecture.

\section*{Acknowledgement}
We thank Taekyung Kim and Namuk Park for comments on the self-supervised pre-training.
This work was supported by an IITP grant funded by the Korean Government (MSIT) (RS-2020-II201361, Artificial Intelligence Graduate School Program (Yonsei University)) and by the Yonsei Signature Research Cluster Program of 2024 (2024-22-0161).

\input{appendix}

{
    \footnotesize
    \renewcommand{\bibname}{\protect\leftline{References}}
    \renewcommand\bibpreamble{\vspace{-3\baselineskip}}
    \setlength{\bibsep}{0pt}
    \bibliographystyle{ieeenat_fullname}
    \bibliography{main}
}

\end{document}

%% file: preamble.tex
\usepackage[dvipsnames]{xcolor}

\usepackage{algorithm}
\usepackage{algorithmic}
\usepackage{epsfig}
\usepackage{graphicx}
\usepackage{amsmath}
\usepackage{amssymb}

\usepackage{url}
\usepackage{graphicx}
\usepackage{subcaption}
\usepackage{booktabs}
\usepackage{tabularx}
\usepackage{multirow}

\usepackage{xspace}
\usepackage{newfloat}
\usepackage{listings}
\DeclareCaptionStyle{ruled}{labelfont=normalfont,labelsep=colon,strut=off} %
\floatstyle{ruled}
\newfloat{listing}{tb}{lst}{}
\floatname{listing}{Listing}

\usepackage{xspace}
\newcommand{\oursfull}{Similarity by Attack Transferability (SAT)\xspace}
\newcommand{\ours}{SAT\xspace}
\newcommand{\ourss}{SATs\xspace}

\definecolor{bestcolor}{RGB}{252, 229, 205}
\newcommand{\best}[1]{\cellcolor{bestcolor}{\textbf{#1}}}
\definecolor{secondcolor}{RGB}{243, 243, 243}

%% file: appendix.tex
\appendix
\numberwithin{equation}{section}
\numberwithin{figure}{section}
\numberwithin{table}{section}

\section*{Author Contributions}
This work is done as an internship project by J Hwang under the supervision of S Chun. S Chun initialized the project idea: understanding how different architectures behave differently by using an adversarial attack. J Hwang, S Chun, and D Han jointly designed the analysis toolbox. J Hwang implemented the analysis toolbox and conducted the experiments with input from S Chun, D Han, and J Lee. J Hwang, D Han, B Heo, and S Chun contributed to interpreting and understanding various neural architectures under our toolbox. The initial version of ``model card'' (\cref{tab:feature-1} and \ref{tab:features-2}) was built by J Hwang, S Park, and verified by D Han and B Heo. B Heo contributed to interpreting distillation results. All ResNet and ViT models newly trained in this work were trained by S Park. J Lee supervised J Hwang and verified the main idea and experiments during the project. S Chun and J Hwang wrote the initial version of the manuscript. All authors contributed to the final manuscript.

\section{Empirical Comparison of SAT and Other Methods}
\label{sec:appendix_empricial_comparison}

\begin{table}
\footnotesize
\centering
\caption{\small{\bf Comparison of stability of measurements.} We compare the stability of method by \cite{somepalli2022reproducibility} and SAT. Stability is indicated by the standard deviation (std). The numbers in ($\cdot$) mean the sampling ratio to all possible combinations to compute the exact value. ``cost'' denotes relative costs compared to the total forward costs for the 50K ImageNet validation set: Somepalli \etal needs $50\text{K} \choose 3$ = 2.1$\times 10^{13}$ and \ours needs 50K. Here, we assume that forward and backward computations cost the same.}
\label{tab:some_comparison}
\begin{tabular}{cccccc}
\toprule
\multicolumn{3}{c}{ \citet{somepalli2022reproducibility}}&\multicolumn{3}{c}{\ours (ours)}\\
\# triplets&std&cost&\# images&std&cost\\
\midrule
10 (4.8$\times 10^{-13}$)&4.49&1.0&500 (0.01)&1.88&1.0\\
20 (9.6$\times 10^{-13}$)&3.28&2.0&1000 (0.02)&1.05&2.0\\
50 (2.4$\times 10^{-12}$)&1.63&5.0&2500 (0.05)&0.91&5.2\\
100 (4.8$\times 10^{-12}$)&1.54&10.0&5000 (0.1)&0.77&10.2\\
\bottomrule
\end{tabular}
\end{table}

\subsection{Comparison with \citeauthor{somepalli2022reproducibility} in the Variance of Similarity}
\label{subsec:appendix_empricial_comparison_somepalli}

\citet{somepalli2022reproducibility} proposed a sampling-based similarity score for comparing decision boundaries of models. \ours has two advantages over Somepalli \etal: computational efficiency and the reliability of the results. First, \ours involves sampling 5,000 images and using 50-step PGD; the computation cost is [5,000 (sampled images) $\times$ 50 (PGD steps) + 5,000 (test to the other model)] $\times$ 2 (two models). 
Meanwhile, \citet{somepalli2022reproducibility} sample 500 triplets and generate 2,500 points to construct decision boundaries. In this case, the total inference cost is [500 (sampled triplets) $\times$ 2,500 (grid points)] $\times$ 2 (two models), which is 4.9 times larger than SAT.
Secondly, Somepalli \etal sampled three images of different classes. As the original paper used CIFAR-10 \cite{cifar10}, 500 triplets can cover all possible combinations of three classes among the ten classes ($10 \choose 3$ = 120 $<$ 500). However, it becomes computationally infeasible to represent all possible combinations of three classes among many classes (\eg, ImageNet needs $1000 \choose 3$ = 164,335,500, 130 times greater than its training images).
Also, we find that the similarity of \cite{somepalli2022reproducibility} is unreliable 
when the number of sample triplets is small. In \cref{tab:some_comparison}, we calculate the similarity scores between \texttt{ConvNeXt-T} \cite{convnext} and \texttt{Swin-T} \cite{swin} from ten different sets with varying sample sizes. \ours exhibits significantly better stability (\ie, low variances) than Somepalli \etal. Note that we use our similarity measurement as the percentage degree without the logarithmic function and control the scale of samples to maintain similar computation complexity between \ours and the compared method for a fair comparison.

\subsection{Ensemble Performance Comparison of \ours and Other Methods}
\label{subsec:appendix_empricial_comparison_ensemble}

The purpose of our study is to provide a new lens for model similarity through adversarial attack transferability. Since we do not have the ground truth of the ``similarity'' between architectures, comparing different similarity functions is not really meaningful. Instead, we indirectly compare \ours, \citet{somepalli2022reproducibility} and naive architecture feature-based clustering on the ensemble benchmark. More specifically, the naive architecture clustering is based on our architecture features proposed in \cref{subsec:model-analysis}; we apply the K-means clustering algorithm to get clusters. We note that the other comparison methods cannot be applied due to the expensive computations. The results are shown in \cref{tab:appendix_ens1} and \cref{tab:appendix_ens2}.

\begin{table}[h!]
\begin{minipage}[b]{0.43\linewidth}
\scriptsize
\caption{Somepalli \etal}
\label{tab:appendix_ens1}
\renewcommand{\arraystretch}{0.9}
\resizebox{\textwidth}{!}{
\setlength{\tabcolsep}{1.5pt}
\begin{tabular}{cc|ccccc}
& & \multicolumn{5}{c}{\tiny less diverse $\leftarrow$ \# of clusters $\rightarrow$ more diverse}\\
& & 1 & 2 & 3 & 4 & 5 \\
\midrule
\parbox[t]{2mm}{\multirow{4}{*}{\rotatebox[origin=c]{90}{{\tiny \# of models}}}} &2 & 7.54 & \textbf{7.79} & & & \\
&3 & 10.85 & 11.03 & \textbf{11.13} & & \\
&4 & 12.63 & 12.79 & 12.87 & \textbf{12.93} & \\
&5 & 13.74 & 13.89 & 13.95 & 14.01 & \textbf{14.05}\\
\end{tabular}
}
\end{minipage}
\begin{minipage}[b]{0.55\linewidth}
\scriptsize
\caption{Arc-based clustering}
\label{tab:appendix_ens2}
\renewcommand{\arraystretch}{0.9}
\setlength{\tabcolsep}{1.5pt}
\resizebox{\textwidth}{!}{
\begin{tabular}{cc|ccccc|cc}
& & \multicolumn{5}{c|}{\tiny less diverse $\leftarrow$ \# of clusters $\rightarrow$ more diverse}& \\
& & 1 & 2 & 3 & 4 & 5 & rand & SAT \\
\midrule
\parbox[t]{2mm}{\multirow{4}{*}{\rotatebox[origin=c]{90}{{\tiny \# of models}}}}&2 & 7.60 & \textbf{7.80} & & & & 7.78 & \best{7.84} \\
&3 & 11.02 & 11.08 & \textbf{11.12} & & & 11.11 & \best{11.20}\\
&4 & 13.03 & 12.92 & 12.89 & 12.90 & & 12.90 & \best{13.00} \\
&5 & 14.37 & 14.13 & 14.03 & 14.00 & 14.00 & 14.01 & \best{14.11}\\
\end{tabular}
}
\end{minipage}
\end{table}

In the tables, SAT is the best-performing model similarity score on the ensemble task.
We also tried to \citet{tramer2017space} in the same setting, but \citet{tramer2017space} often failed to converge and show very small differences. Hence, we couldn't use \citet{tramer2017space} for measuring all 69 arches used in the paper.
Instead, as we reported in the main paper, we compare \citet{tramer2017space} and other model similarity variants on 8 architectures used in \cref{fig:pgd-otherattack}. Among the candidate methods, we observe that \ours shows the strongest correlation with the ensemble performance using two models.

\section{Discussions}
\label{sec:appendix_discussions}

\subsection{Robustness of SAT to the choice of the attack methods.}
\label{subsec:appendix_attack_robustness}

\begin{figure}[t!]
    \centering
    \begin{subfigure}[b]{0.32\linewidth}
    \includegraphics[width=\linewidth]{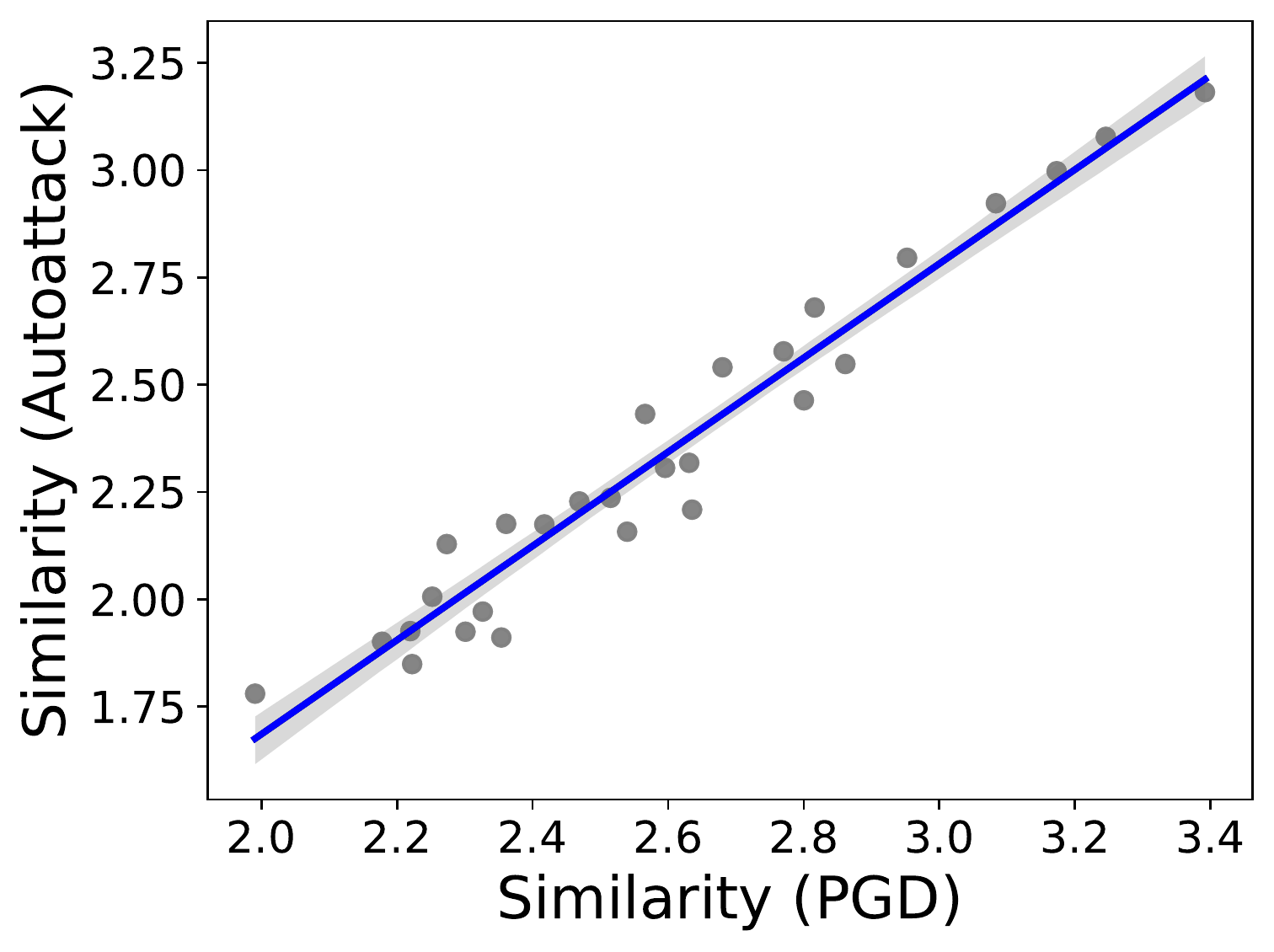}
    \caption{ PGD vs. Autoattack}
    \label{fig:pgd-autoattack}
    \end{subfigure}
    \begin{subfigure}[b]{0.32\linewidth}
    \includegraphics[width=\linewidth]{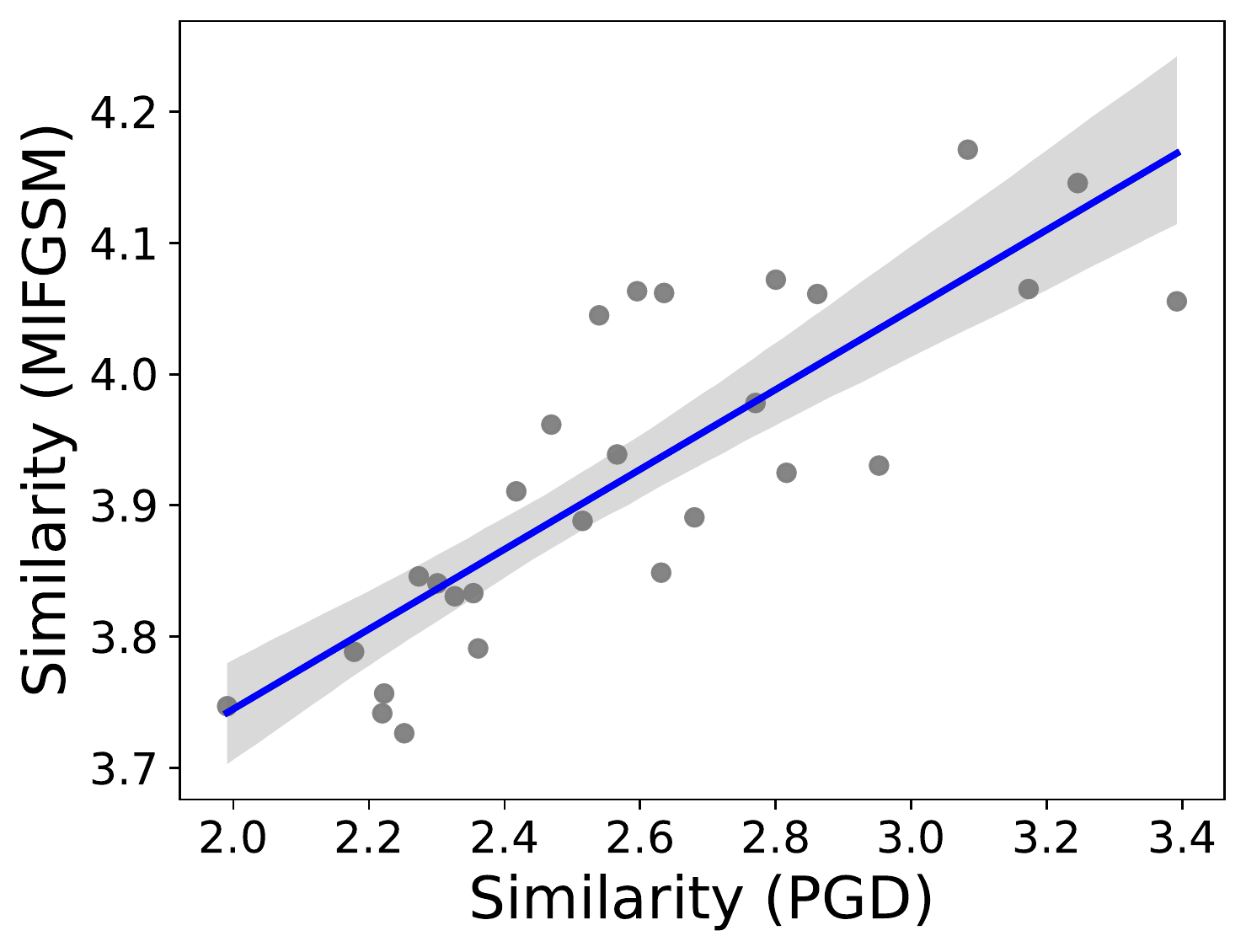}
    \caption{ PGD vs. MIFGSM}
    \label{fig:pgd-mifgsm}
    \end{subfigure}
    \begin{subfigure}[b]{0.32\linewidth}
    \includegraphics[width=\linewidth]{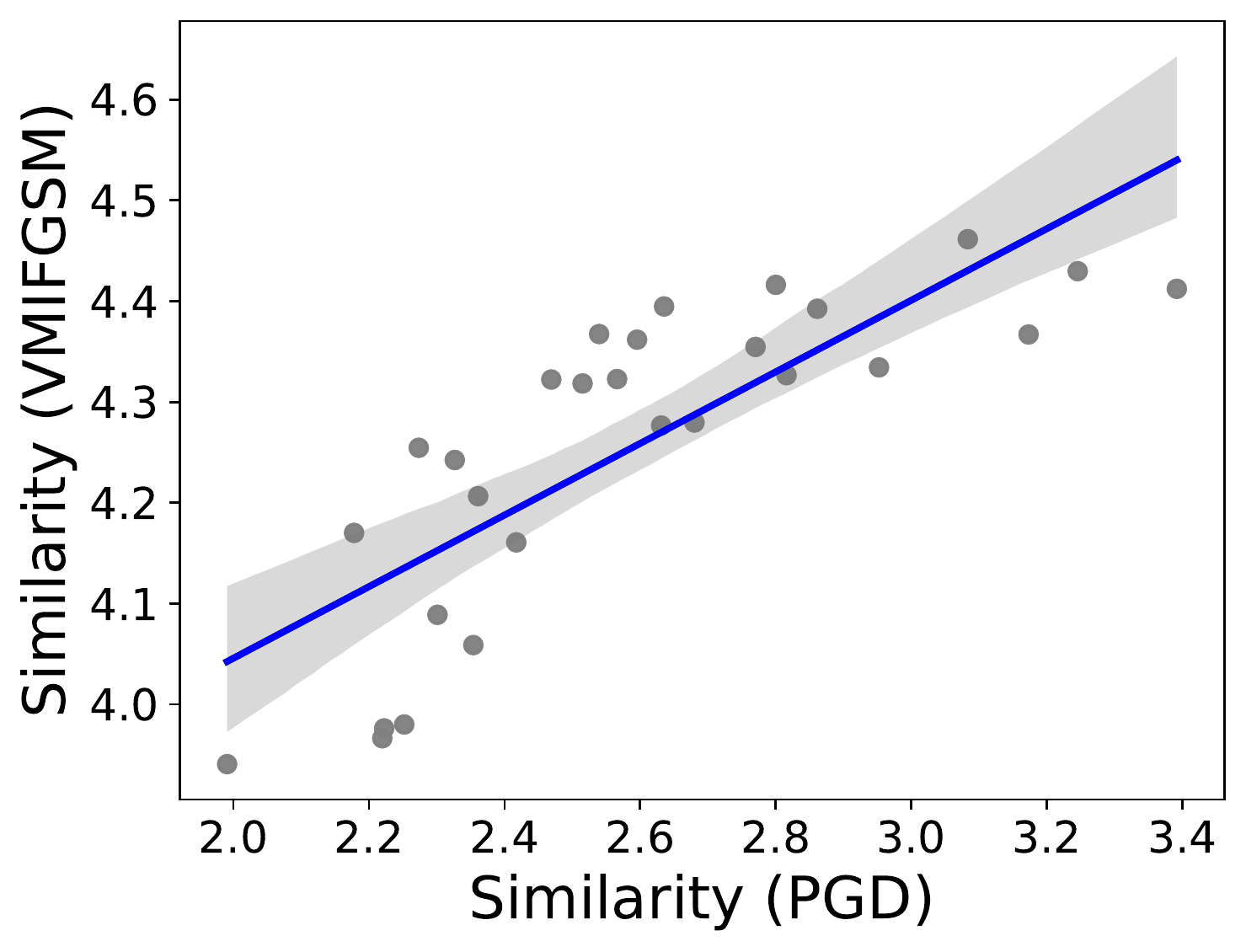}
    \caption{ PGD vs. VMIFGSM}
    \label{fig:pgd-vmifgsm}
    \end{subfigure}
    \begin{subfigure}[b]{0.32\linewidth}
    \includegraphics[width=\linewidth]{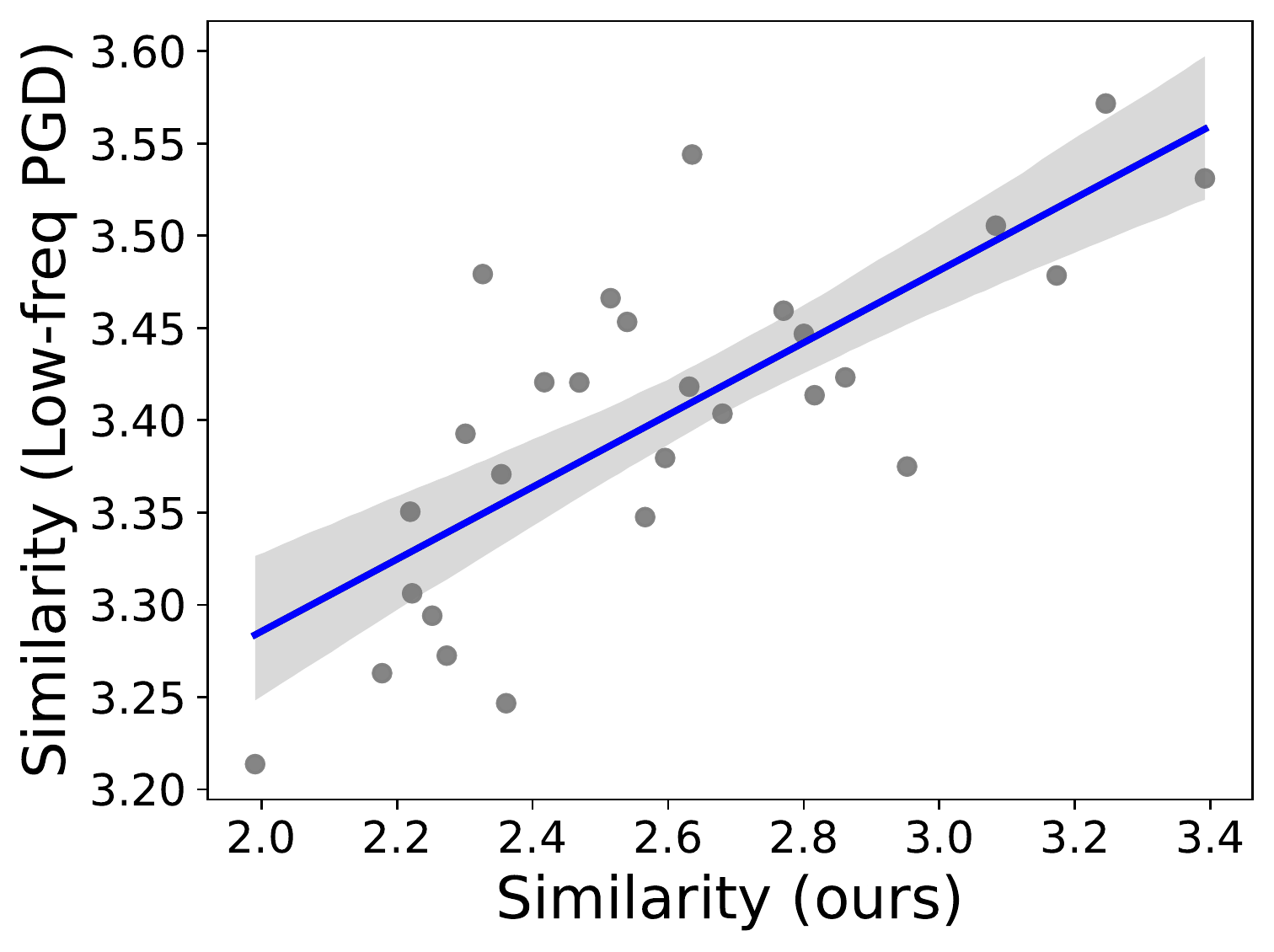}
    \caption{PGD vs. Low Freq. PGD}
    \label{fig:pgd-bia}
    \end{subfigure}
    \begin{subfigure}[b]{0.32\linewidth}
    \includegraphics[width=\linewidth]{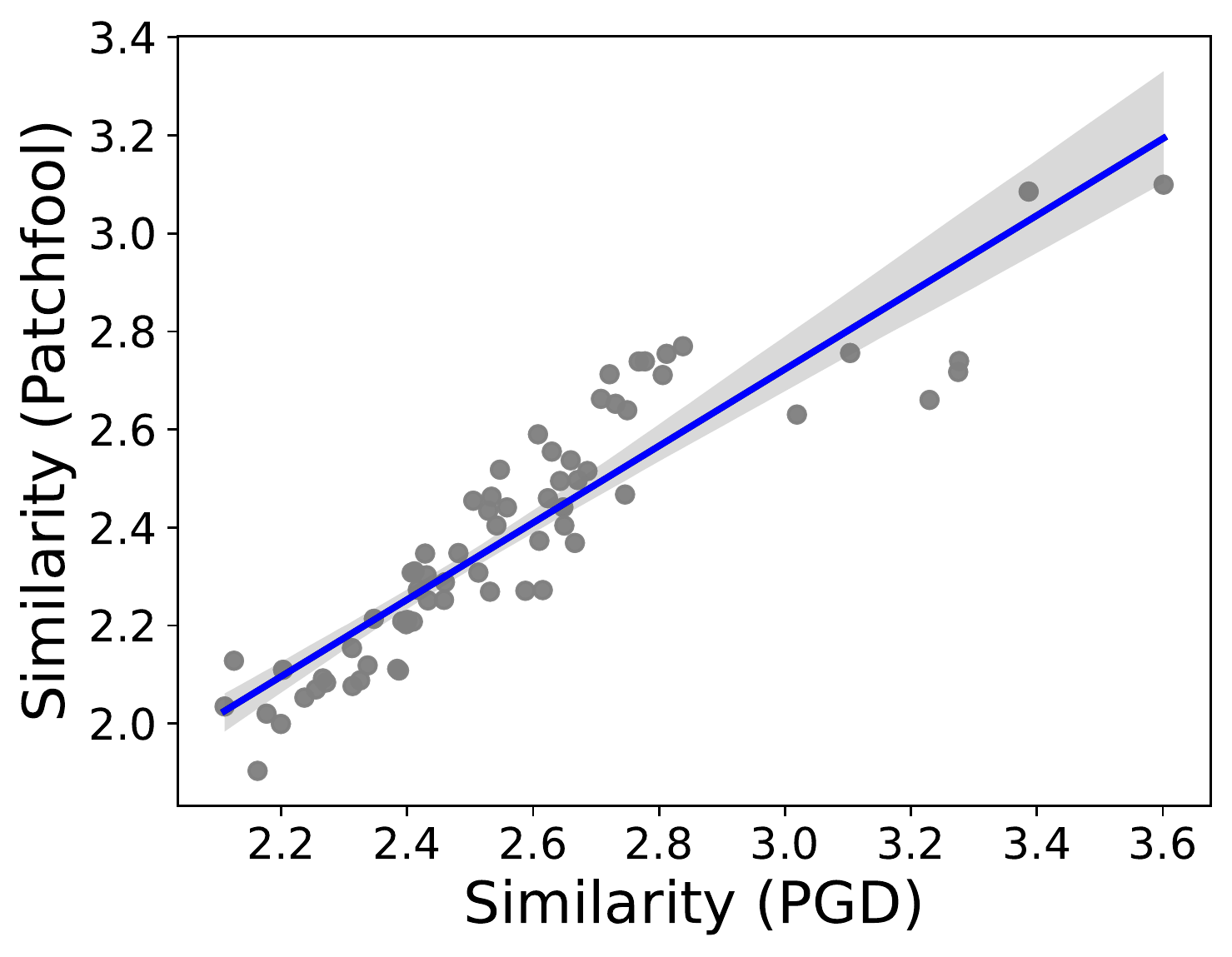}
    \caption{ PGD vs. Patchfool}
    \label{fig:pgd-patchfool}
    \end{subfigure}
    \begin{subfigure}[b]{0.32\linewidth}
    \includegraphics[width=\linewidth]{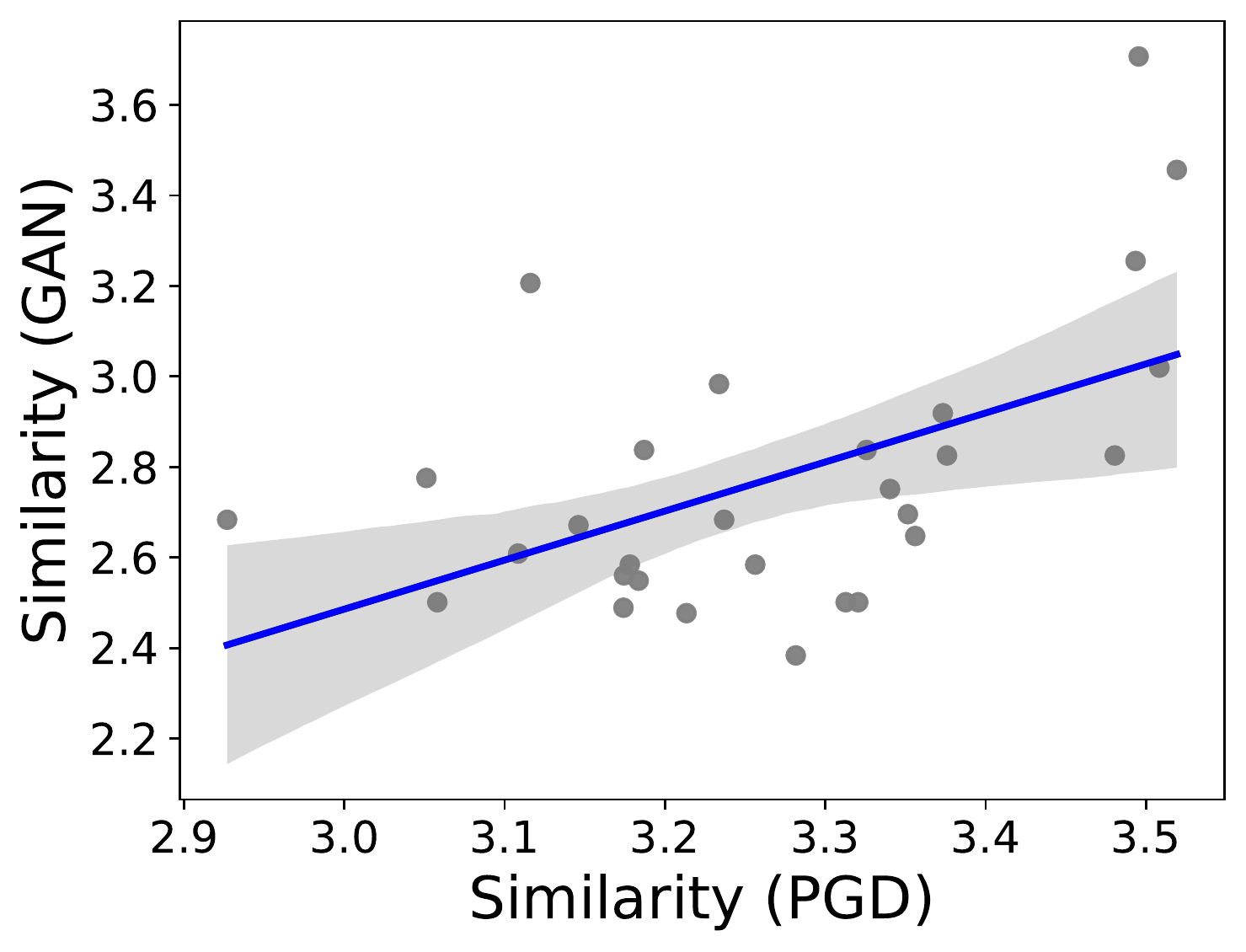}
    \caption{PGD vs. BIA}
    \label{fig:pgd-bia}
    \end{subfigure}
    \caption{\small {\bf Effect of different adversarial attack methods to \ours.} The trend line and its 90\% confidence interval are shown. We show the relationship between our \ours using PGD \cite{madry2017pgd} and \ours using other attacks, (a) Autoattack \cite{autoattack} (b) MIFGSM \cite{MIFGSM}(c) VMIFGSM \cite{VMIFGSM} (d) low-frequency PGD \cite{lowfpgd} (e) Patchfool attack \cite{patchfool}, and (f) BIA \cite{BIA}.
    }
    \label{fig:pgd-otherattack}
\end{figure}

Our goal is not to design an attack-free method but to show the potential of using adversarial attack transferability (AAT) for measuring quantitative model similarity. However, we have checked that SAT scores are robust to the choice of the attack methods, even for stronger attacks, such as AutoAttack \cite{autoattack}, attacks designed for enhancing AAT, such as MIFGSM \cite{MIFGSM} and VMIFGSM \cite{VMIFGSM}, low-frequency targeted attacks, such as low-frequency PGD \cite{lowfpgd}, method-specific attacks, such as PatchPool \cite{patchfool}, or generative model-based attacks, such as BIA \cite{BIA}.
We sample 8 representative models among 69 models for testing the effect of attacks on \ours;
\texttt{ViT-S}, \texttt{CoaT-Lite Small}, \texttt{ResNet-101}, \texttt{LamHaloBotNet50}, \texttt{ReXNet-150}, \texttt{NFNet-L0}, \texttt{Swin-T} and \texttt{Twins-pcpvt}. 
\cref{fig:pgd-autoattack} shows the high correlation between \ours scores using PGD and Autoattack; it shows a correlation coefficient of 0.98 with a p-value of $1.43 \times 10^{-18}$. For testing the Patchfool attack, we only generate adversarial perturbations on \texttt{ViT-S} and get attack transferability to all other models (68 models) because it only targets Transformers. 
\cref{fig:pgd-mifgsm} and \ref{fig:pgd-vmifgsm} show similar results: SAT shows consistent results even for the attacks designed for better AAT. The results show that \ours score is robust to the choice of attack methods if the attack is strong enough.
Also, the low-frequency attack \cite{lowfpgd} shows a similar result, i.e., the frequency-targeted attack does not affect the similarity results.
In \cref{fig:pgd-patchfool}, Patchfool also shows a high correlation compared to the PGD attack (correlation coefficient 0.91 with p-value $3.62 \times 10^{-27}$). 
We additionally provide a result for generative model-based attack, BIA \cite{BIA}. As BIA needs to train a new generative model for a different architecture, we only show the pre-trained models provided by the authors. We also get a similar result with previous results for BIA.
Note that the compared attack methods are not model agnostic or computationally expensive than PGD, \eg, PatchFool needs a heavy modification on the model code to extract attention layer outputs manually, and BIA needs to train a new generator for a new architecture. As SAT shows consistent rankings across the attack methods, we use PGD due to its simplicity.

\subsection{Impact of Adversarial Training to \ours}
\label{subsec:appendix_impact_of_adv_tr}
While our main analyses are based on ImageNet-trained models, in this subsection, we use CIFAR-10-trained models for two reasons. First, it is still challenging to achieve a high-performing adversarially trained model on the ImageNet scale. On the other hand, in the CIFAR-10 training setting, a number of adversarially-trained models are available and comparable. Second, adversarial training models show lower clean accuracy than normally trained models \cite{tsipras2018robustness}. Adversarial robustness and accuracy are in a trade-off, and there is no ImageNet model with accuracy aligned with our target models yet.

We choose five adversarial training ResNet-18 from the AutoAttack repository \cite{autoattack} and measure \ours using the models. The average \ours between adversarial training models is \textbf{3.15}, slightly lower than the similarity score with different training strategies for ImageNet ResNet-50 (3.27 -- See \cref{tab:inner-similarity}). In other words, we can confirm that different adversarial training methods make as a difference as different training techniques.

\section{Details of Architectures Used in the Analyses}
\label{sec:appendix-models}

We use 69 models in our research to evaluate the similarity between models and to investigate the impact of model diversity. In the main paper, we mark the names of models based on their research paper and PyTorch Image Models library (\texttt{timm}; 0.6.7 version) \citep{rw2019timm}. \cref{tab:list} shows the full list of the models based on their research paper and \texttt{timm} alias.

We show brief information of the architectural components in \cref{tab:model-features}. The full network specification is shown in \cref{tab:feature-1} and \cref{tab:features-2}.
We follow the corresponding paper and \texttt{timm} library to list the model features. 

\begin{table}[h!]
\caption{\textbf{Lists of 69 models and their names based on their research paper and \texttt{timm} library.}}
\label{tab:list}
\resizebox{\textwidth}{!} {
\setlength{\tabcolsep}{4pt}
\begin{tabular}{ll|ll|ll}
\toprule
in \texttt{timm} &in paper& in \texttt{timm}&in paper& in \texttt{timm}&in paper\\
\midrule
botnet26t\_256& \texttt{BoTNet-26} & gluon\_xception65& \texttt{Xception-65}& resnet50\_gn & \texttt{ResNet-50 (GN)}\\
coat\_lite\_small & \texttt{CoaT-Lite Small} & gmlp\_s16\_224 & \texttt{gMLP-S} & resnetblur50 & \texttt{ResNet-50} (BlurPool)\\
convit\_base& \texttt{ConViT-B}& halo2botnet50ts\_256 & \texttt{Halo2BoTNet-50} & resnetv2\_101& \texttt{ResNet-V2-101} \\
convmixer\_1536\_20 & \texttt{ConvMixer-1536/20} & halonet50ts& \texttt{HaloNet-50} & resnetv2\_50 & \texttt{ResNet-V2-50}\\
convnext\_tiny& \texttt{ConvNeXt-T}& haloregnetz\_b & \texttt{HaloRegNetZ}& resnetv2\_50d\_evos& \texttt{ResNet-V2-50-EVOS} \\
crossvit\_base\_240 & \texttt{CrossViT-B}& hrnet\_w64 & \texttt{HRNet-W32}& resnext50\_32x4d & \texttt{ResNeXt-50}\\
cspdarknet53& \texttt{CSPDarkNet-53} & jx\_nest\_tiny & \texttt{NesT-T} & rexnet\_150& \texttt{ReXNet ($\times$1.5)} \\
cspresnet50 & \texttt{CSPResNet-50}& lambda\_resnet50ts & \texttt{LambdaResNet-50}& sebotnet33ts\_256& \texttt{SEBoTNet-33} \\
cspresnext50& \texttt{CSPResNeXt-50} & lamhalobotnet50ts\_256 & \texttt{LamHaloBoTNet-50} & sehalonet33ts& \texttt{SEHaloNet-33}\\
deit\_base\_patch16\_224& \texttt{DeiT-B}& mixnet\_xl & \texttt{MixNet-XL}& seresnet50 & \texttt{SEResNet-50} \\
deit\_small\_patch16\_224 & \texttt{DeiT-S}& nf\_regnet\_b1 & \texttt{NF-RegNet-B1} & seresnext50\_32x4d & \texttt{SEResNeXt-50}\\
dla102x2& \texttt{DLA-X-102} & nf\_resnet50 & \texttt{NF-ResNet-50} & swin\_s3\_tiny\_224& \texttt{S3} (Swin-T) \\
dpn107& \texttt{DPN-107} & nfnet\_l0& \texttt{NFNet-L0} & swin\_tiny\_patch4\_window7\_224 & \texttt{Swin-T}\\
eca\_botnext26ts\_256 & \texttt{ECA-BoTNeXt-26}& pit\_b\_224& \texttt{PiT-B}& tnt\_s\_patch16\_224 & \texttt{TNT-S} \\
eca\_halonext26ts & \texttt{ECA-HaloNeXt-26} & pit\_s\_224& \texttt{PiT-S}& twins\_pcpvt\_base & \texttt{Twins-PCPVT-B} \\
eca\_nfnet\_l0& \texttt{ECA-NFNet-L0}& poolformer\_m48& \texttt{PoolFormer-M48} & twins\_svt\_small& \texttt{Twins-SVT-S} \\
eca\_resnet33ts & \texttt{ECA-ResNet-33} & regnetx\_320 & \texttt{RegNetX-320}& visformer\_small & \texttt{VisFormer-S} \\
efficientnet\_b2& \texttt{EfficientNet-B2} & regnety\_032 & \texttt{RegNetY-32} & vit\_base\_patch32\_224& \texttt{ViT-B} \\
ese\_vovnet39b& \texttt{eSE-VoVNet-39} & resmlp\_24\_224& \texttt{ResMLP-S24} & vit\_small\_patch16\_224 & \texttt{ViT-S} \\
fbnetv3\_g& \texttt{FBNetV3-G} & resmlp\_big\_24\_224 & \texttt{ResMLP-B24} & vit\_small\_r26\_s32\_224& \texttt{R26+ViT-S} \\
gcresnet50t & \texttt{GCResNet-50} & resnest50d & \texttt{ResNeSt-50} & wide\_resnet50\_2& \texttt{Wide ResNet-50}\\
gcresnext50ts & \texttt{GCResNeXt-50}& resnet101& \texttt{ResNet-101} & xcit\_medium\_24\_p16\_224 & \texttt{XCiT-M24}\\
gluon\_resnet101\_v1c & \texttt{ResNet-101-C}& resnet50 & \texttt{ResNet-50}& xcit\_tiny\_12\_p8\_224& \texttt{XCiT-T12}\\
\bottomrule
\end{tabular}
}
\end{table}
\begin{table}[h!]
\small
\centering
\caption{\small {\bf Overview of model elements.} We categorize each architecture with 13 different architectural components. The full feature list is in the Appendix.
}
\label{tab:model-features}
\resizebox{\textwidth}{!} {
\begin{tabular}{@{}rl@{}}
\toprule
Components & Elements \\ \midrule
Base architecture & CNN, Transformer, MLP-Mixer, Hybrid (CNN + Transformer), NAS-Net \\
Stem layer & 7$\times$7 conv with stride 2, 3$\times$3 conv with stride 2, 16$\times$16 conv with stride 16, ...\\
Input resolution & 224$\times$224, 256$\times$256, 240$\times$240, 299$\times$299 \\
Normalization layer & BN, GN, LN, LN + GN, LN + BN, Normalization-free, ... \\
Using hierarchical structure & Yes (\eg, CNNs, Swin \cite{swin}), No (\eg, ViT \cite{vit}) \\
Activation functions & ReLU, HardSwish, SiLU, GeLU, ReLU + GeLU, ReLU + SiLU or GeLU  ... \\
Using pooling at stem  & Yes, No \\
Using 2D self-attention & Yes (\eg, \cite{lambdanet}, \cite{botnet}, \cite{halonet}), No \\
Using channel-wise (CW) attention & Yes (\eg, \cite{senet}, \cite{gcnet}, \cite{ecanet}), No \\
Using depth-wise convolution & Yes, No \\
Using group convolution & Yes, No \\
Type of pooling for final feature & Classification (CLS) token, Global Average Pool (GAP) \\
Location of CW attentions & At the end of each block, in the middle of each block, ... \\
 \bottomrule
\end{tabular}
}
\end{table}

\begin{table}[htp!]
\centering
\footnotesize
\caption{\small Description of features of 69 models. ``s" in ``Stem layer" indicates the stride of a layer in the stem, and the number before and after ``s" are a kernel size and size of stride, respectively. For example, ``3s2/3/3" means that the stem is composed of the first layer having 3 $\times$ 3 kernel with stride 2, the second layer having 3 $\times$ 3 kernel with stride 1, and the last layer having 3 $\times$ 3 with stride 1.}
\label{tab:feature-1}
\resizebox{\textwidth}{!} {
\setlength{\tabcolsep}{8pt}
\begin{tabular}{c||c|c|c|c|c|c}
\toprule
Model name&Base architecture&\shortstack[lb]{Hierarchical\\structure}&Stem layer &Input resolution&Normalization&Activation\\
\midrule
botnet26t\_256 & CNN & Yes & 3s2/3/3 & 256 $\times$ 256 & BN & ReLU  \\ \hline
convmixer\_1536\_20 & CNN & Yes & 7s7 & 224 $\times$ 224 & BN & GeLU  \\ \hline
convnext\_tiny & CNN & Yes & 4s4 & 224 $\times$ 224 & LN & GeLU  \\ \hline
cspdarknet53 & CNN & Yes & 3s1 & 256 $\times$ 256 & BN & Leaky ReLU  \\ \hline
cspresnet50 & CNN & Yes & 7s2 & 256 $\times$ 256 & BN & Leaky ReLU  \\ \hline
cspresnext50 & CNN & Yes & 7s2 & 256 $\times$ 256 & BN & Leaky ReLU  \\ \hline
dla102x2 & CNN & Yes & 7s1 & 224 $\times$ 224 & BN & ReLU  \\ \hline
dpn107 & CNN & Yes & 7s2 & 224 $\times$ 224 & BN & ReLU  \\ \hline
eca\_botnext26ts\_256 & CNN & Yes & 3s2/3/3 & 256 $\times$ 256 & BN & SiLU  \\ \hline
eca\_halonext26ts & CNN & Yes & 3s2/3/3 & 256 $\times$ 256 & BN & SiLU  \\ \hline
eca\_nfnet\_l0 & CNN & Yes & 3s2/3/3/3s2 & 224 $\times$ 224 & Norm-free & SiLU  \\ \hline
eca\_resnet33ts & CNN & Yes & 3s2/3/3s2 & 256 $\times$ 256 & BN & SiLU  \\ \hline
ese\_vovnet39b & CNN & Yes & 3s2/3/3s2 & 224 $\times$ 224 & BN & ReLU  \\ \hline
gcresnet50t & CNN & Yes & 3s2/3/3s2 & 256 $\times$ 256 & LN + BN & ReLU  \\ \hline
gcresnext50ts & CNN & Yes & 3s2/3/3 & 256 $\times$ 256 & LN + BN & ReLU + SiLU  \\ \hline
gluon\_resnet101\_v1c & CNN & Yes & 3s2/3/3 & 224 $\times$ 224 & BN & ReLU  \\ \hline
gluon\_xception65 & CNN & Yes & 3s2/3 & 299 $\times$ 299 & BN & ReLU  \\ \hline
halo2botnet50ts\_256 & CNN & Yes & 3s2/3/3s2 & 256 $\times$ 256 & BN & SiLU  \\ \hline
halonet50ts & CNN & Yes & 3s2/3/3 & 256 $\times$ 256 & BN & SiLU  \\ \hline
hrnet\_w64 & CNN & Yes & 3s2/3s2 & 224 $\times$ 224 & BN & ReLU \\ \hline
lambda\_resnet50ts & CNN & Yes & 3s2/3/3 & 256 $\times$ 256 & BN & SiLU  \\ \hline
lamhalobotnet50ts\_256 & CNN & Yes & 3s2/3/3s2 & 256 $\times$ 256 & BN & SiLU  \\ \hline
nf\_resnet50 & CNN & Yes & 7s2 & 256 $\times$ 256 & Norm-free & ReLU  \\ \hline
nfnet\_l0 & CNN & Yes & 3s2/3/3/3s2 & 224 $\times$ 224 & Norm-free & ReLU + SiLU  \\ \hline
poolformer\_m48 & CNN & Yes & 7s4 & 224 $\times$ 224 & GN & GeLU  \\ \hline
resnest50d & CNN & Yes & 3s2/3/3 & 224 $\times$ 224 & BN & ReLU  \\ \hline
resnet101 & CNN & Yes & 7s2 & 224 $\times$ 224 & BN & ReLU  \\ \hline
resnet50 & CNN & Yes & 7s2 & 224 $\times$ 224 & BN & ReLU  \\ \hline
resnet50\_gn & CNN & Yes & 7s2 & 224 $\times$ 224 & GN & ReLU  \\ \hline
resnetblur50 & CNN & Yes & 7s2 & 224 $\times$ 224 & BN & ReLU  \\ \hline
resnetv2\_101 & CNN & Yes & 7s2 & 224 $\times$ 224 & BN & ReLU  \\ \hline
resnetv2\_50 & CNN & Yes & 7s2 & 224 $\times$ 224 & BN & ReLU  \\ \hline
resnetv2\_50d\_evos & CNN & Yes & 3s2/3/3 & 224 $\times$ 224 & EvoNorm & -  \\ \hline
resnext50\_32x4d & CNN & Yes & 7s2 & 224 $\times$ 224 & BN & ReLU  \\ \hline
sebotnet33ts\_256 & CNN & Yes & 3s2/3/3s2 & 256 $\times$ 256 & BN & ReLU + SiLU  \\ \hline
sehalonet33ts & CNN & Yes & 3s2/3/3s2 & 256 $\times$ 256 & BN & ReLU + SiLU  \\ \hline
seresnet50 & CNN & Yes & 7s2 & 224 $\times$ 224 & BN & ReLU  \\ \hline
seresnext50\_32x4d & CNN & Yes & 7s2 & 224 $\times$ 224 & BN & ReLU  \\ \hline
wide\_resnet50\_2 & CNN & Yes & 7s2 & 224 $\times$ 224 & BN & ReLU \\ \hline
convit\_base & Transformer & No & 16s16 & 224 $\times$ 224 & LN & GeLU  \\ \hline
crossvit\_base\_240 & Transformer & Yes & 16s16 & 240 $\times$ 240 & LN & GeLU  \\ \hline
deit\_base\_patch16\_224 & Transformer & No & 16s16 & 224 $\times$ 224 & LN & GeLU  \\ \hline
deit\_small\_patch16\_224 & Transformer & No & 16s16 & 224 $\times$ 224 & LN & GeLU  \\ \hline
jx\_nest\_tiny & Transformer & Yes & 4s4 & 224 $\times$ 224 & LN & GeLU  \\ \hline
pit\_s\_224 & Transformer & Yes & 16s8 & 224 $\times$ 224 & LN & GeLU  \\ \hline
swin\_tiny\_patch4\_window7\_224 & Transformer & Yes & 4s4 & 224 $\times$ 224 & LN & GeLU  \\ \hline
tnt\_s\_patch16\_224 & Transformer & Yes & 7s4 & 224 $\times$ 224 & LN & GeLU  \\ \hline
vit\_base\_patch32\_224 & Transformer & No & 32s32 & 224 $\times$ 224 & LN & GeLU  \\ \hline
vit\_small\_patch16\_224 & Transformer & No & 16s16 & 224 $\times$ 224 & LN & GeLU \\ \hline
gmlp\_s16\_224 & MLP-Mixer & Yes & 16s16 & 224 $\times$ 224 & LN & GeLU  \\ \hline
resmlp\_24\_224 & MLP-Mixer & No & 16s16 & 224 $\times$ 224 & Affine transform & GeLU  \\ \hline
resmlp\_big\_24\_224 & MLP-Mixer & Yes & 8s8 & 224 $\times$ 224 & Affine transform & GeLU \\ \hline
swin\_s3\_tiny\_224 & NAS (TFM) & Yes & 4s4 & 224 $\times$ 224 & LN & GeLU  \\ \hline
efficientnet\_b2 & NAS (CNN) & Yes & 3s2 & 256 $\times$ 256 & BN & SiLU  \\ \hline
fbnetv3\_g & NAS (CNN) & Yes & 3s2 & 240 $\times$ 240 & BN & HardSwish  \\ \hline
haloregnetz\_b & NAS (CNN) & Yes & 3s2 & 224 $\times$ 224 & BN & ReLU + SiLU  \\ \hline
mixnet\_xl & NAS (CNN) & Yes & 3s2 & 224 $\times$ 224 & BN & ReLU + SiLU  \\ \hline
nf\_regnet\_b1 & NAS (CNN) & Yes & 3s2 & 256 $\times$ 256 & Norm-free & ReLU + SiLU  \\ \hline
regnetx\_320 & NAS (CNN) & Yes & 3s2 & 224 $\times$ 224 & BN & ReLU  \\ \hline
regnety\_032 & NAS (CNN) & Yes & 3s2 & 224 $\times$ 224 & BN & ReLU  \\ \hline
rexnet\_150 & NAS (CNN) & Yes & 3s2 & 224 $\times$ 224 & BN & ReLU + SiLU + ReLU6 \\ \hline
coat\_lite\_small & Hybrid & Yes & 4s4 & 224 $\times$ 224 & LN & GeLU  \\ \hline
pit\_b\_224 & Hybrid & Yes & 14s7 & 224 $\times$ 224 & LN & GeLU  \\ \hline
twins\_pcpvt\_base & Hybrid & Yes & 4s4 & 224 $\times$ 224 & LN & GeLU  \\ \hline
twins\_svt\_small & Hybrid & Yes & 4s4 & 224 $\times$ 224 & LN & GeLU  \\ \hline
visformer\_small & Hybrid & Yes & 7s2 & 224 $\times$ 224 & BN & GeLU + ReLU  \\ \hline
vit\_small\_r26\_s32\_224 & Hybrid & No & 7s2 & 224 $\times$ 224 & LN + GN & GeLU + ReLU  \\ \hline
xcit\_medium\_24\_p16\_224 & Hybrid & No & 3s2/3s2/3s2/3s2 & 224 $\times$ 224 & LN + BN & GeLU  \\ \hline
xcit\_tiny\_12\_p8\_224 & Hybrid & No & 3s2/3s2/3s2 & 224 $\times$ 224 & LN + BN & GeLU \\ \hline
\end{tabular}
}
\end{table}
\clearpage

\begin{table}[htp!]
\centering
\footnotesize
\caption{\small Description of features of 69 models. ``Pooling (stem)" and ``Pooling (final)" denote ``Pooling at the stem" and ``Pooling for final feature", respectively. ``SA", ``CW", and ``DW" means ``Self-attention", ``Channel-wise", and ``Depth-wise", respectively.}
\label{tab:features-2}
\resizebox{\textwidth}{!} {
\setlength{\tabcolsep}{7pt}
\begin{tabular}{c||c|c|c|c|c|c|c}
\toprule
Model name&Pooling (stem)&Pooling (final)&2D SA&CW attention&\shortstack[lb]{Location of\\CW attention}&DW conv&Group conv\\
\midrule
botnet26t\_256 & Yes & GAP & Yes (BoT) & No & ~ & No & No  \\ \hline
convmixer\_1536\_20 & No & GAP & No & No & ~ & Yes & No  \\ \hline
convnext\_tiny & No & GAP & No & No & ~ & Yes & No  \\ \hline
cspdarknet53 & No & GAP & No & No & ~ & No & No  \\ \hline
cspresnet50 & Yes & GAP & No & No & ~ & No & No  \\ \hline
cspresnext50 & Yes & GAP & No & No & ~ & No & Yes  \\ \hline
dla102x2 & No & GAP & No & No & ~ & No & Yes  \\ \hline
dpn107 & Yes & GAP & No & No & ~ & No & Yes  \\ \hline
eca\_botnext26ts\_256 & Yes & GAP & Yes (BoT) & Yes (ECA) & Middle & No & Yes  \\ \hline
eca\_halonext26ts & Yes & GAP & Yes (Halo) & Yes (ECA) & Middle & No & Yes  \\ \hline
eca\_nfnet\_l0 & No & GAP & No & Yes (ECA) & End & No & Yes  \\ \hline
eca\_resnet33ts & No & GAP & No & Yes (ECA) & Middle & No & No  \\ \hline
ese\_vovnet39b & No & GAP & No & Yes (ESE) & End & No & No  \\ \hline
gcresnet50t & No & GAP & No & Yes (GCA) & Middle & Yes & No  \\ \hline
gcresnext50ts & Yes & GAP & No & Yes (GCA) & Middle & Yes & Yes  \\ \hline
gluon\_resnet101\_v1c & Yes & GAP & No & No & ~ & No & No  \\ \hline
gluon\_xception65 & No & GAP & No & No & ~ & Yes & No  \\ \hline
halo2botnet50ts\_256 & No & GAP & Yes (Halo, BoT) & No & ~ & No & No  \\ \hline
halonet50ts & Yes & GAP & Yes (Halo) & No & ~ & No & No  \\ \hline
hrnet\_w64 & No & GAP & No & No & ~ & No & No \\ \hline
lambda\_resnet50ts & Yes & GAP & Yes (Lambda) & No & ~ & No & No  \\ \hline
lamhalobotnet50ts\_256 & No & GAP & Yes (Lambda, Halo, BoT) & No & ~ & No & No  \\ \hline
nf\_resnet50 & Yes & GAP & No & No & ~ & No & No  \\ \hline
nfnet\_l0 & No & GAP & No & Yes (SE) & End & No & Yes  \\ \hline
poolformer\_m48 & No & GAP & No & No & ~ & No & No  \\ \hline
resnest50d & Yes & GAP & No & Yes & ~ & No & Yes  \\ \hline
resnet101 & Yes & GAP & No & No & ~ & No & No  \\ \hline
resnet50 & Yes & GAP & No & No & ~ & No & No  \\ \hline
resnet50\_gn & Yes & GAP & No & No & ~ & No & No  \\ \hline
resnetblur50 & Yes & GAP & No & No & ~ & No & No  \\ \hline
resnetv2\_101 & Yes & GAP & No & No & ~ & No & No  \\ \hline
resnetv2\_50 & Yes & GAP & No & No & ~ & No & No  \\ \hline
resnetv2\_50d\_evos & Yes & GAP & No & No & ~ & No & No  \\ \hline
resnext50\_32x4d & Yes & GAP & No & No & ~ & No & Yes  \\ \hline
sebotnet33ts\_256 & No & GAP & Yes (BoT) & Yes (SE) & Middle & No & No  \\ \hline
sehalonet33ts & No & GAP & Yes (Halo) & Yes (SE) & Middle & No & No  \\ \hline
seresnet50 & Yes & GAP & No & Yes (SE) & End & No & No  \\ \hline
seresnext50\_32x4d & Yes & GAP & No & Yes (SE) & End & No & Yes  \\ \hline
wide\_resnet50\_2 & Yes & GAP & No & No & ~ & No & No \\ \hline
convit\_base & No & CLS token & No & No & ~ & No & No  \\ \hline
crossvit\_base\_240 & No & CLS token & No & No & ~ & No & No  \\ \hline
deit\_base\_patch16\_224 & No & CLS token & No & No & ~ & No & No  \\ \hline
deit\_small\_patch16\_224 & No & CLS token & No & No & ~ & No & No  \\ \hline
jx\_nest\_tiny & No & GAP & No & No & ~ & No & No  \\ \hline
pit\_s\_224 & No & CLS token & No & No & ~ & Yes & No  \\ \hline
swin\_tiny\_patch4\_window7\_224 & No & GAP & No & No & ~ & No & No  \\ \hline
tnt\_s\_patch16\_224 & No & CLS token & No & No & ~ & No & No  \\ \hline
vit\_base\_patch32\_224 & No & CLS token & No & No & ~ & No & No  \\ \hline
vit\_small\_patch16\_224 & No & CLS token & No & No & ~ & No & No \\ \hline
gmlp\_s16\_224 & No & GAP & No & No & ~ & No & No  \\ \hline
resmlp\_24\_224 & No & GAP & No & No & ~ & No & No  \\ \hline
resmlp\_big\_24\_224 & No & GAP & No & No & ~ & No & No \\ \hline
swin\_s3\_tiny\_224 & No & GAP & No & No & ~ & No & No  \\ \hline
efficientnet\_b2 & No & GAP & No & Yes (SE) & Middle & Yes & No  \\ \hline
fbnetv3\_g & No & GAP & No & Yes (SE) & Middle & Yes & No  \\ \hline
haloregnetz\_b & No & GAP & Yes (Halo) & Yes (SE) & Middle & No & Yes  \\ \hline
mixnet\_xl & No & GAP & No & Yes (SE) & Middle & Yes & No  \\ \hline
nf\_regnet\_b1 & No & GAP & No & Yes (SE) & Middle & Yes & Yes  \\ \hline
regnetx\_320 & No & GAP & No & No & ~ & No & Yes  \\ \hline
regnety\_032 & No & GAP & No & Yes (SE) & Middle & No & Yes  \\ \hline
rexnet\_150 & No & GAP & No & Yes (SE) & Middle & Yes & No \\ \hline
coat\_lite\_small & No & CLS token & No & No & ~ & Yes & No  \\ \hline
pit\_b\_224 & No & CLS token & No & No & ~ & Yes & No  \\ \hline
twins\_pcpvt\_base & No & GAP & No & No & ~ & Yes & No  \\ \hline
twins\_svt\_small & No & GAP & No & No & ~ & Yes & No  \\ \hline
visformer\_small & No & GAP & No & No & ~ & No & Yes  \\ \hline
vit\_small\_r26\_s32\_224 & Yes & CLS token & No & No & ~ & No & No  \\ \hline
xcit\_medium\_24\_p16\_224 & No & CLS token & No & No & ~ & Yes & No  \\ \hline
xcit\_tiny\_12\_p8\_224 & No & CLS token & No & No & ~ & Yes & No \\ \hline
\end{tabular}
}
\end{table}
\clearpage

\section{Feature Importance and Clustering Details}
\label{sec:appendix_feature_importance_and_clustering}

\subsection{Feature Importance Analysis Details}
\label{subsec:appendix_feature_importance_details}

We fit a gradient boosting regressor \citep{friedman2001greedy} based on \texttt{Scikit-learn} \citep{pedregosa2011scikit} and report the permutation importance of each architectural component in Fig. 3 of the main paper. The number of boosting stages, maximum depth, minimum number of samples, and learning rate are set to 500, 12, 4, and 0.02, respectively.
Permutation importance is computed by permuting a feature 10 times.

\begin{figure}[h!]
    \centering
    \includegraphics[width=\linewidth]{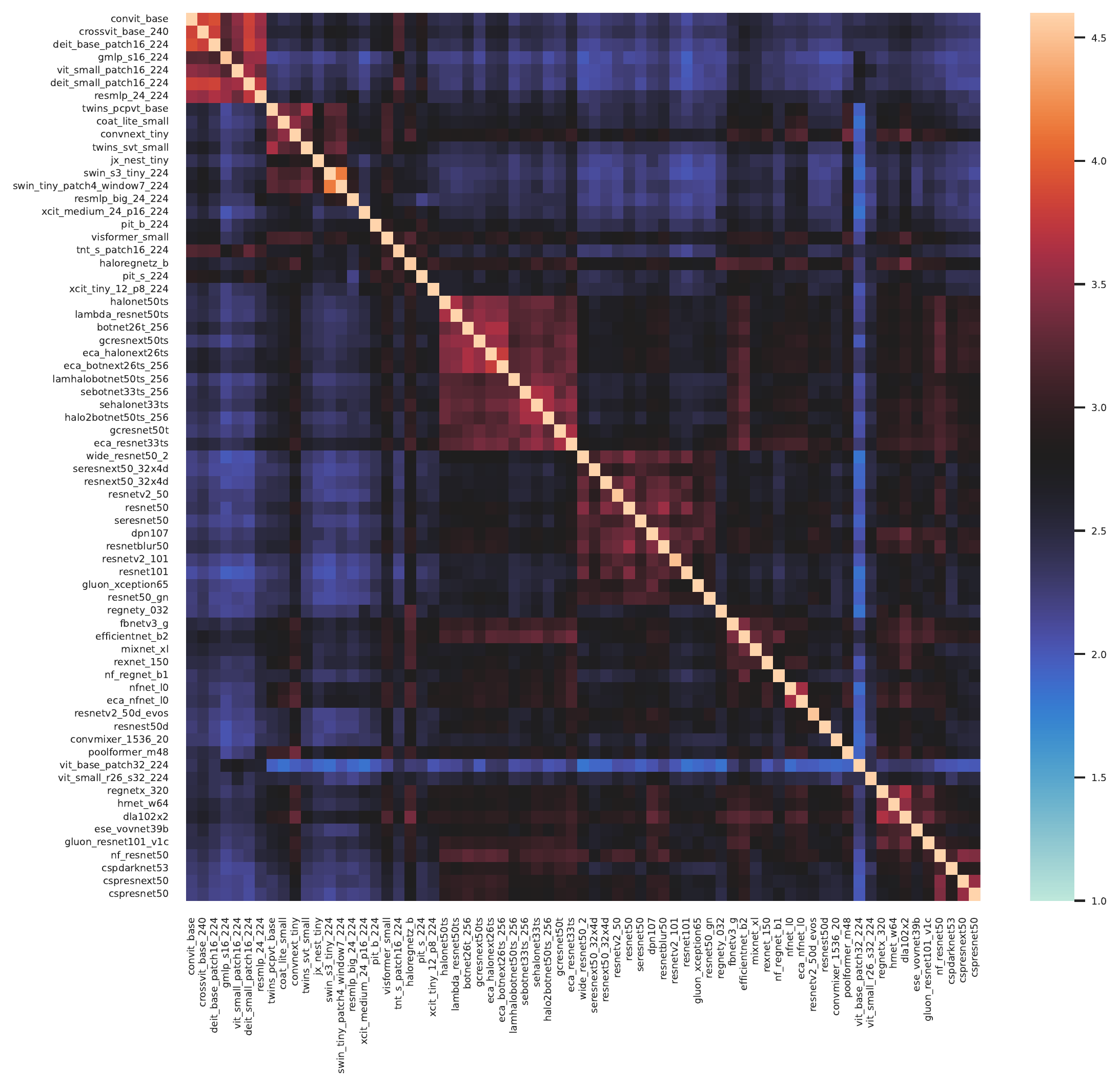}
    \caption{\small {\bf Pairwise similarity among 69 models.} Rows and columns are sorted by the clustering index in Tab. 2. ($n$, $n$) component of pairwise similarity is close to 4.6 ($\log100$) because the attack success rate is almost 100\% when a model used for generating adversarial perturbation and attacked model are the same.}
    \label{fig:pairwise-similarity}
\end{figure}

\subsection{Pairwise Similarities}
\label{subsec:appendix-similarities}

\cref{fig:pairwise-similarity} indicates the pairwise similarity among 69 models. We can observe a weak block pattern around clusters, as also revealed in \cref{fig:model-clustering-blocks}. 

\subsection{Spectral Clustering Details}
\label{subsec:appendix-spectral-clustering}

We use the normalized Laplacian matrix to compute the Laplacian matrix. We also run K-means clustering 100 times and choose the clustering result with the best final objective function to reduce the randomness by the K-means clustering.

\begin{figure}[h!]
    \centering
    \includegraphics[width=\linewidth]{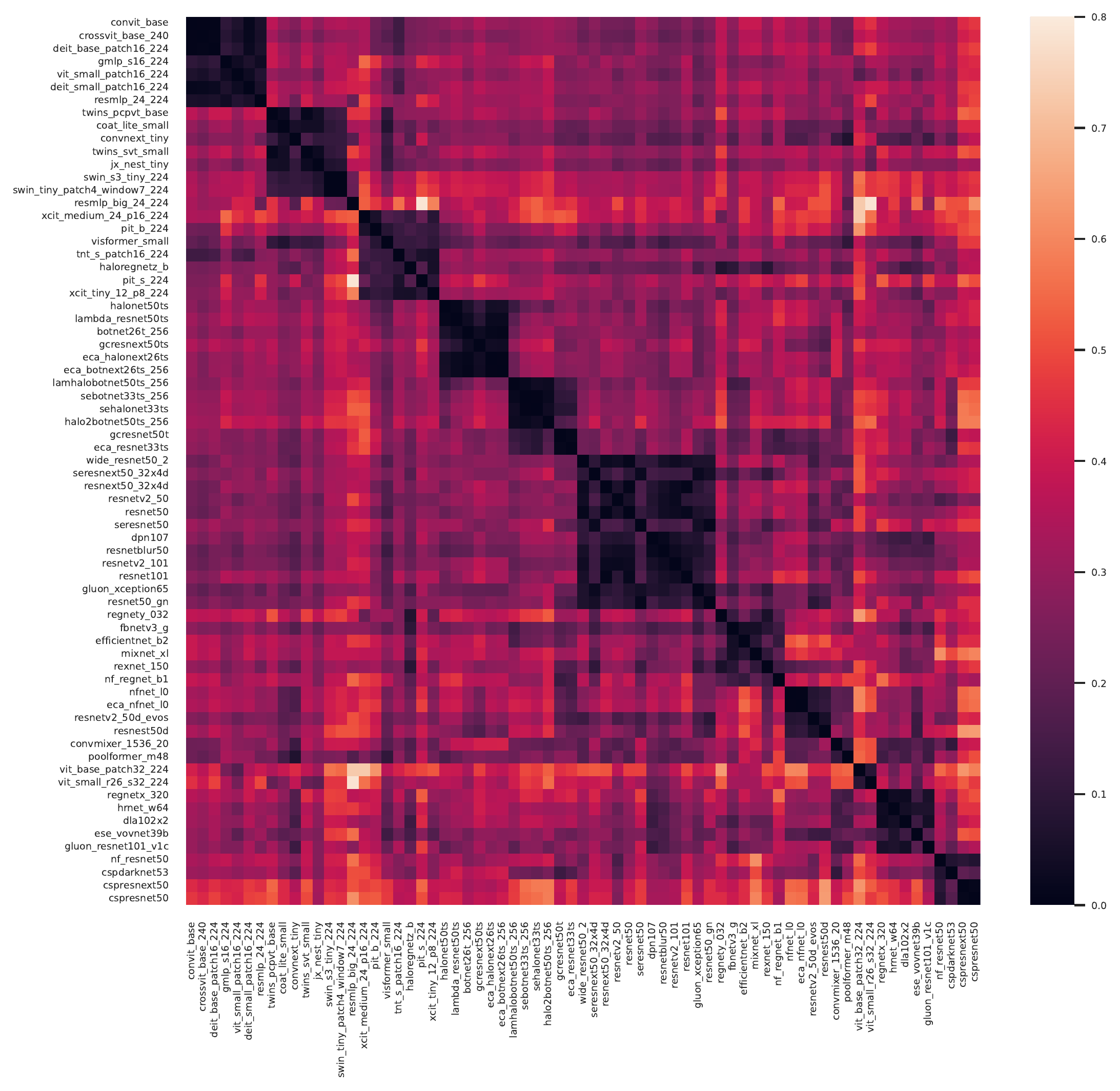}
    \caption{\small {\bf Spectral features of 69 architectures.} The $K$-th largest eigenvectors of the Laplacian matrix of the pairwise similarity graph of 69 architectures are shown ($K=10$ in this figure). Rows and columns are sorted by the clustering index in Tab. 2. We denote the model name in \texttt{timm} for each row and column.}
    \label{fig:model-clustering-blocks}
\end{figure}

We visualize the pairwise distances of the spectral features (\ie, $K$-largest eigenvectors of $L$) of 69 architectures with their model names in \cref{fig:model-clustering-blocks}. This figure is an extension of \cref{fig:similarity-different-vs-same}, now including the model names.
Note that rows and columns of \cref{fig:model-clustering-blocks} are sorted by the clustering results. \cref{fig:model-clustering-blocks} shows block diagonal patterns, \ie, in-cluster similarities are large while between-cluster similarities are small.

\section{Training Settings for Models Used in Analyses}
\label{sec:appendix-different-training-settings}

\paragraph{Models with various training methods for \cref{subsec:model-analysis}.}
We train 21 \texttt{ResNet-50} models and 16 \texttt{ViT-S} from scratch individually by initializing each network with different random seeds.
We further train 28 \texttt{ResNet-50} models by randomly choosing learning rate ($\times$0.1, $\times$0.2, $\times$0.5, $\times$1, $\times$2, and $\times$5 where the base learning rate is 0.1), weight decay ($\times$0.1, $\times$0.2, $\times$0.5, $\times$1, $\times$2, and $\times$5 where the base weight decay is 1e-4), and learning rate scheduler (step decay or cosine decay).
Similarly, we train 9 \texttt{ViT-S} models by randomly choosing learning rate ($\times$0.2, $\times$0.4, and $\times$1 where the base learning rate is 5e-4) and weight decay ($\times$0.2, $\times$0.4, and $\times$1 where the base weight decay is 0.05). Note that the DeiT training is unstable when we use a larger learning rate or weight decay than the base values.
Finally, we collect 22 \texttt{ResNet-50} models with different training regimes: 1 model with standard training by PyTorch \citep{pytorch}; 4 models trained by GluonCV \citep{gluoncv}\footnote{\scriptsize gluon\_resnet50\_v1b, gluon\_resnet50\_v1c, gluon\_resnet50\_v1d, and gluon\_resnet50\_v1s from \texttt{timm} library.}; a semi-supervised model and semi-weakly supervised model on billion-scale unlabeled images by Yalniz \etal \cite{yalniz2019billion}\footnote{\scriptsize ssl\_resnet50 and swsl\_resnet50 from \texttt{timm} library.}; 5 models trained by different augmentation methods (Cutout \citep{devries2017cutout}, Mixup \citep{zhang2017mixup}, manifold Mixup \citep{verma2019manifold}, CutMix \citep{yun2019cutmix}, and feature CutMix\footnote{\scriptsize We use the official weights provided by \url{https://github.com/clovaai/CutMix-PyTorch}.}; 10 optimized ResNet models by \citep{rsb}\footnote{ \scriptsize We use the official weights provided by \url{https://github.com/rwightman/pytorch-image-models/releases/tag/v0.1-rsb-weights}}.
We also collect 7 \texttt{ViT-S} models with different training regimes, including the original ViT training setup~\citep{vit}\footnotemark{\value{footnote}-3}, a stronger data augmentation setup in the Deit paper~\citep{deit}\footnotemark[\value{footnote}-3], the training setup with distillation~\citep{deit}\footnotemark[\value{footnote}-3], an improved DeiT training setup~\citep{deit3}\footnotemark[\value{footnote}-3] \footnotetext{\scriptsize deit\_small\_patch16\_224, vit\_small\_patch16\_224, deit\_small\_distilled\_patch16\_224, and deit3\_small\_patch16\_224 from \texttt{timm} library.}
, and self-supervised training fashions by MoCo v3 \citep{mocov3}\footnote{\scriptsize We train the \texttt{ViT-S} model by following \url{https://github.com/facebookresearch/moco-v3}}, MAE \citep{mae}\footnote{\scriptsize We train the \texttt{ViT-S} model by following \url{https://github.com/facebookresearch/mae}} and BYOL \citep{byol}\footnote{\scriptsize We train the \texttt{ViT-S} model by following \url{ https://github.com/lucidrains/byol-pytorch}}. 
We do not use adversarially-trained networks because the adversarial training usually drops the standard accuracy significantly~\cite{tsipras2018robustness}.

\paragraph{Distillation models for \cref{sec:applications}.}
We train \texttt{ViT-Ti} student models with 25 different teacher models using hard distillation strategy.
We follow the distillation training setting of DeiT official repo\footnote{\url{https://github.com/facebookresearch/deit}.}, only changing the teacher model. Note that we resize the input images to the input size the teacher model requires if the input sizes of student and teacher models differ.
If a teacher model needs a different input resolution, such as $240\times240$ and $256\times256$, we resize the input image for distilling it.
Because \texttt{DeiT-Ti} has low classification accuracy compared to teacher models, the similarity score is calculated between \texttt{DeiT-S} and 25 models.
The 25 teacher models are as follows: {\small \texttt{BoTNet-26}, \texttt{CoaT-Lite Small}, \texttt{ConViT-B}, \texttt{ConvNeXt-B}, \texttt{CrossViT-B}, \texttt{CSPDarkNet-53}, \texttt{CSPResNeXt-50}, \texttt{DLA-X-102}, \texttt{DPN-107}, \texttt{EfficientNet-B2}, \texttt{FBNetV3-G}, \texttt{GC-ResNet-50}, \texttt{gMLP-S}, \texttt{HaloRegNetZ}, \texttt{MixNet-XL}, \texttt{NFNet-L0}, \texttt{PiT-S}, \texttt{RegNetY-032}, \texttt{ResMLP-24}, \texttt{ResNet-50}, \texttt{SEHaloNet33}, \texttt{Swin-T}, \texttt{TNT-S}, \texttt{VisFormer-S}, and \texttt{XCiT-T12}}. 

\section{Knowledge Distillation}

\subsection{Teacher Accuracy and Distillation Performance}
\label{subsec:appendix-distill-teacher}

\begin{figure}[h]
    \centering
    \includegraphics[width=0.6\linewidth]{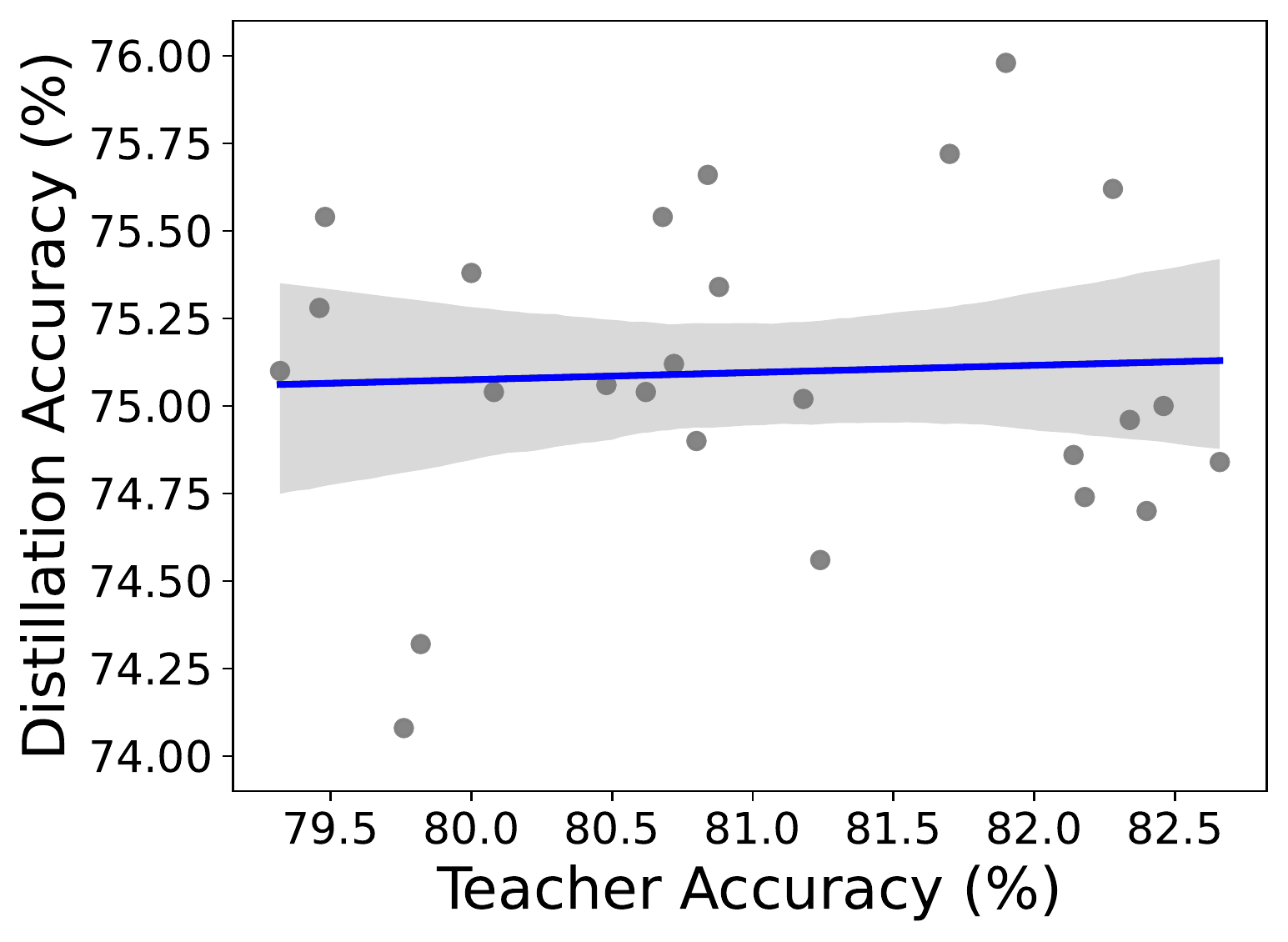}%
	\caption{\small {\bf Teacher network accuracy and distillation performance.}
    There is no significant correlation between teacher accuracy and distillation performance.}
	\label{fig:distill-teacher}
\end{figure}
The similarity between teacher and student networks may not be the only factor contributing to distillation. For example, a stronger teacher can lead to better distillation performance \citep{yun2021relabel}.
In \cref{fig:distill-teacher}, we observe that if the range of the teacher accuracy is not significantly large enough (\eg, between 79.5 and 82.5), then the correlation between teacher network accuracy and distillation performance is not significant; 0.049 Pearson correlation coefficient with 0.82 p-value.
In this case, we can conclude that the teacher and student networks similarity contributes more to the distillation performance than the teacher performance.

\section{Limtations and Discussions}
\label{sec:appendix-limitations-discussions}

\subsection{Efficient Approximation of SAT for a Novel Model}
\label{subsec:efficient_approx}

\begin{figure}[t]
    \centering
    \begin{subfigure}[b]{0.49\linewidth}
    \includegraphics[width=\linewidth]{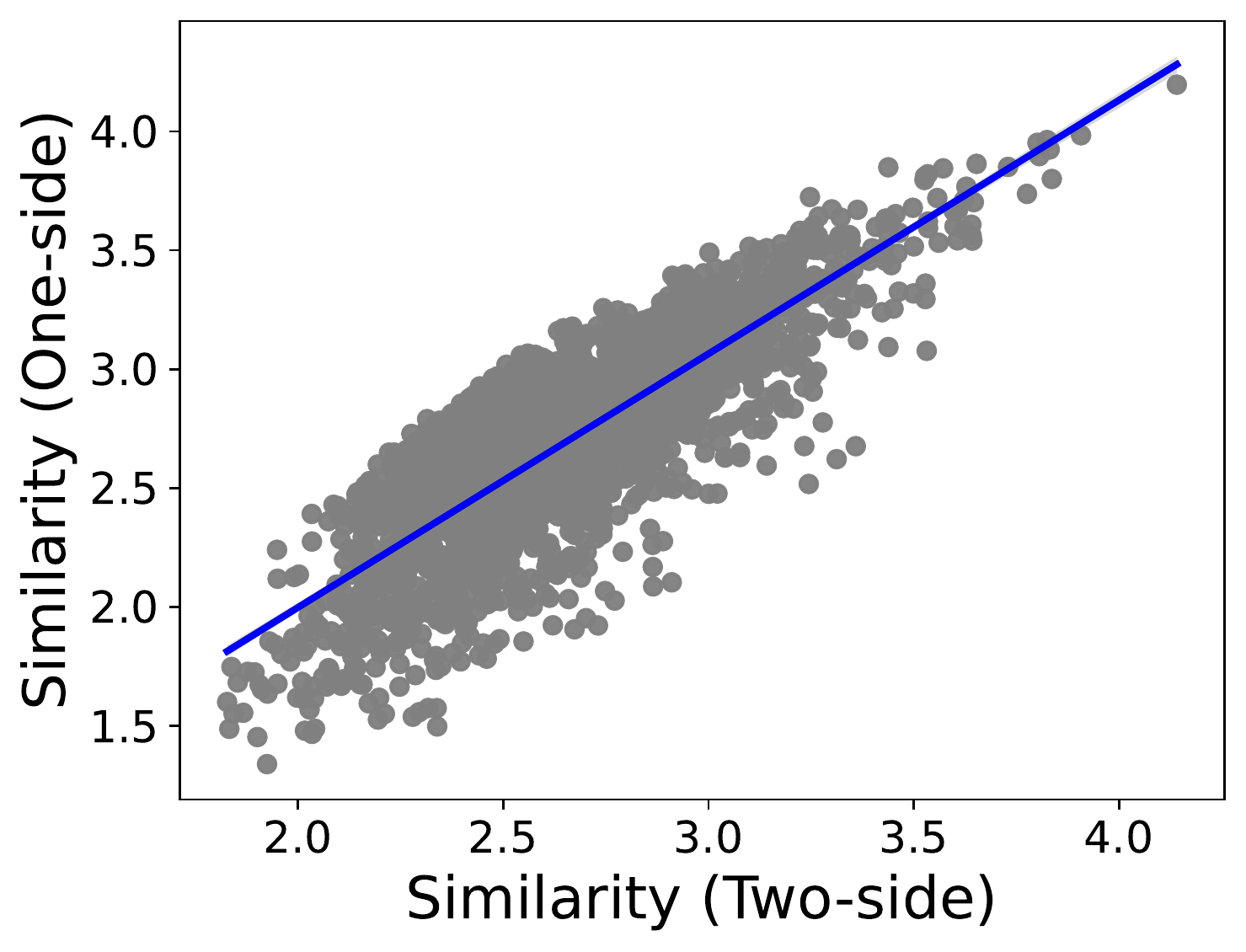}
    \caption{\small {\bf Approximated (one-side) SAT vs. original (two-side).}}
	\label{fig:one-two-side}
    \end{subfigure}
    \begin{subfigure}[b]{0.49\linewidth}
    \includegraphics[width=\linewidth]{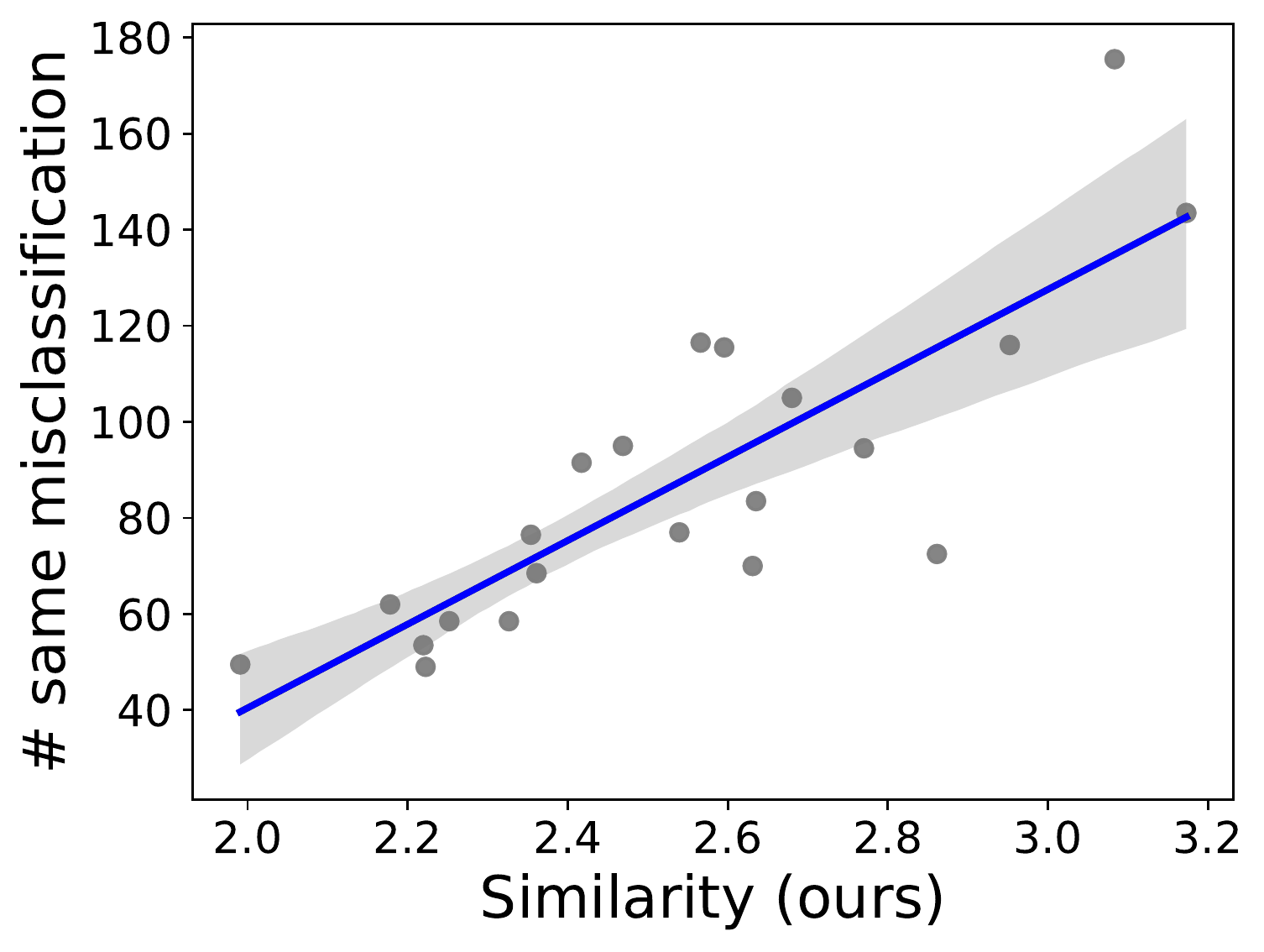}
    \caption{\small \bf SAT vs. The number of same misclassification.}
    \label{fig:where}
    \end{subfigure}
    \caption{\small {\bf Additional Analysis for SAT.}}
\end{figure}

We can use our toolbox for designing a new model; we can measure SAT between a novel network and existing $N$ architectures; a novel network can be assigned to clusters (\cref{tab:clustering-results}) to understand how it works. However, it requires generating adversarial samples for all $N+1$ models (\eg, 70 in our case), which is computationally inefficient. Instead, we propose the approximation of Eq. \ref{eq:score} by omitting to compute the accuracy of the novel network on the adversarial samples of the existing networks. It will break the symmetricity of SAT, but we found that the approximated score and the original score have high similarity -- 0.82 Pearson coefficient with almost 0 p-value -- as shown in \cref{fig:one-two-side}.

As an example, we tested \texttt{Gluon-ResNeXt-50} \citep{gluoncv} and the distilled version of \texttt{DeiT-S} \citep{deit}. As observed in \cref{tab:inner-similarity} and \cref{fig:similarity-different-vs-same-b}, models with the same architecture have high similarity compared to models with different architectures; hence, we expect that \texttt{Gluon-ResNeXt-50} is assigned to the same cluster with \texttt{ResNeXt-50}, and distilled \texttt{DeiT-S} is assigned to the same cluster with \texttt{DeiT-S}. As we expected, each network is assigned to the desired cluster. Therefore, we suggest using our efficient approximation for analyzing a novel network with our analysis toolbox.

\subsection{Adversarial Attack Transferability and Direction of Missclassification}
\label{subsec:aat_and_misclsf}

\citet{waseda2023closer} showed that adversarial attack transferability is highly related to the direction of the misclassification.
We examine if SAT is related to the misclassification. \cref{fig:where} shows the relationship between SAT and the number of the same misclassification by the attack. We observe that they are highly correlated, \ie, we confirmed that SAT is also related to the misclassification.

\subsection{Change of \ours During Training}
\label{subsec:appendix_sat_during_training}

We check the adversarial attack transferability between the fully trained model and less trained models on CIFAR-10 with 180 training epochs. A model trained with only 20 epochs shows high similarity over different initializations (4.23 in \cref{tab:inner-similarity} of the main paper). Note that \ours considers models having similar clean accuracy; namely, there is room to explore this further in future work.

\begin{table}[h]
\centering
\caption{Change of \ours between fully-trained model (epoch 180) and models on various epochs.}
\begin{tabular}{c|cccccccccc}
\toprule
Epoch& {0}& {20}& {40}& {60}& {80}& {100}& {120}& {140}& {160}& {180}\\
\midrule
{SAT} & {2.84} & {4.46} & {4.56} & {4.58} & {4.59} & {4.59} & {4.60} & {4.60} & {4.60} & {4.60} \\
\bottomrule
\end{tabular}
\end{table}

\subsection{More Possible Applications Requiring Diverse Models}
\label{subsec:appendix_more_apps}
In the main paper, we introduce several applications with multiple models, such as model ensemble, knowledge distillation, and novel model development. As another example, we employ SAT-based diverse model selection for improving the dataset distillation (DD) task with random network selection \cite{zhang2023accelerating}.
DD task \cite{dc,dsa,dcc,zhang2023accelerating,ddreview} aims to synthesize a small (usually less than 5 images per class) but informative dataset that prevents a significant drop from the original performance.
Acc-DD \cite{zhang2023accelerating} employs multiple random networks for DD, where each network is randomly selected during the training.
In this study, we show that a more diverse network selection can help synthesize more informative and diverse condensed images.
We replace the random selection of Acc-DD ({\small \texttt{Rand}}) with the selection by the probability proportional to (1) the similarity ({\small \texttt{P$_\text{sim}$}}) or (2) the inverse of similarity ({\small \texttt{P$_{\text{sim}^{-1}}$}}). More specifically, we first (a) select a network randomly and (b) select the next network by (1) or (2) with the current network. We repeat (b) similar to K-means++ \cite{kmeanspp}. We report the CIFAR-10 results by setting images per class as 1 using 50 CNNs. \texttt{Rand} shows 48.6 top-1 accuracy, while \texttt{P$_\text{sim}$} and \texttt{P$_{\text{sim}^{-1}}$} show 48.7 and \textbf{49.4}, respectively.
Namely, a more diverse network selection (\texttt{P$_{\text{sim}^{-1}}$}) helps Acc-DD.